\documentclass[a4paper,11pt,final]{article}

\usepackage{graphicx,amsmath,amssymb,algorithm,algorithmic} 
\usepackage{epsfig}

\begin{document}

\title{\ \\ \LARGE\bf Visual art inspired by the 
collective
feeding behavior of sand--bubbler crabs}

\author{Hendrik Richter \\
HTWK Leipzig University of Applied Sciences \\ Faculty of
Electrical Engineering and Information Technology\\
        Postfach 301166, D--04251 Leipzig, Germany. \\ Email: 
hendrik.richter@htwk-leipzig.de. }

\maketitle

\begin{abstract} 
Sand--bubblers are crabs of 
 the genera  {\it Dotilla} and {\it Scopimera} which are known to produce remarkable patterns and structures at tropical beaches. From these pattern--making abilities, we may draw inspiration for 
 digital visual art. A simple mathematical model is proposed and an algorithm is designed that may create such sand--bubbler patterns artificially. In addition, design parameters to modify the patterns are identified and analyzed by computational aesthetic measures.  Finally, an extension of the algorithm is discussed that may enable controlling and guiding   generative evolution of the art--making process.

\end{abstract}

\section{Introduction}
Among the various forms of inspiration from biology for digital visual art, swarms and other types of collective animal behavior  are considered to be particularly promising of showing algorithmic ideas and templates that may have the potential to create non--trivial patterns of aesthetic value.  Examples are ant-- and ant--colony--inspired visual art~\cite{green15,urb11}, but our  imagination has also been stimulated by  the pattern--making of flies~\cite{ali17} and other swarms, flocks, or colonies~\cite{jac07,rom08}.   

In this paper, the collective behavior of another family of animals provides a source of inspiration: crabs of  the genera  {\it Dotilla} and {\it Scopimera}, commonly called  sand--bubblers.  Sand--bubbler crabs are known for their
burrow--orientated feeding behavior that produces radial patterns of regular shapes and designs in the wet sand of tropical beaches~\cite{ans88,big21,fied70}.   The patterns consist of tiny balls that are placed in curves or spirals, straight or bent lines, which finally form overall structures, thus producing astonishing works of natural art~\cite{bur16}. The creation of  structures starts at receding tide. The patterns grow in complexity and scale as time goes by until the returning tide cleans them off, in fact resetting the art--making process. This pattern--making behavior, its visual results, and the potentials of  thus inspired algorithms producing digital visual art are our topics.   

We approach these topics by discussing three interconnected main themes in this paper. The first is to describe an aspect of animal behavior observed in nature (the feeding pattern of a colony  of sand--bubbler crabs) by economic mathematical means in order to make it accessible for an algorithmic  process producing two--dimensional visual art. This goes along with identifying design parameters that allow the patterns to vary within the restriction that they should generally resemble structures as found in nature. The second theme is to study the image--producing process by computational aesthetic measures. Here, the focus is not on numerically  evaluating an overall artistic value. There are some  strong arguments that such an evaluation is rather elusive and may not really be achievable free from ambiguity, see also the discussion in Sec. \ref{sec:art}.  Instead, the focus is put on how some aesthetic measures, namely Benford's law measure, Ross, Ralph and Zong's bell--curve measure and  a fractal dimension measure, scale to the design parameters. A third theme is  to use these measures to guide and control the art--making process. This is done by updating the design para\-meters while the algorithmic process is running. By employing the relationships between aesthetic measures and design parameters for updating the parameters, the measures can be changed in an intended way. Hence, this theme studies the possibilities of including generative development, which takes up some ideas from evolving and evolutionary art.

The paper is structured as follows.  Sec. \ref{sec:beh} reviews the behavior of sand--bubbler crabs, particularly the burrow--orientated feeding process.  Inspired by these behavioral patterns, Sec. \ref{sec:art} presents an algorithm to create  visual art, studies computational aesthetic measures and generative development, and finally  shows examples of art pieces thus created. Concluding remarks and an outlook end the paper.

\section{Behavior of sand--bubbler crabs} \label{sec:beh}
Sand--bubbler crabs are  tiny crustacean decapods of the genera  {\it Dotilla} and {\it Scopimera} in the family {\it Dotillidae} that
dwell on tropical sandy beaches across the Indo--Pacific region.  They permanently inhabit intertidal environments as they adapted to the interplay between 
 high and low tide. Sand--bubblers are surface deposit feeders. Thus, their habitat must be periodically covered by water for their food supply to restore. They are also  air--breathers. Thus, the habital zone needs to be periodically uncovered from water for giving them time for feeding.  Observing these periodic changes, the crabs dig burrows in the sand where they seek shelter and hide during high tide. In low tide (and daylight) they swarm out 
in search for food. They pick up sand grains with their pincers and eat the microscopic organic particles that are left over by the receding tide and coat the grains. 
After they have cleaned the sand grains of eatable components, they form them into small balls, also called pseudo--faecal pellets,  and throw them behind. Supposedly, they do so in order 
to avoid searching the same sand twice. As high tide comes back in, they retreat to their burrows, seal them to trap air, and wait for the next tidal cycle.  

\begin{figure}[tb]
\includegraphics[width=5.8cm, height=4.8cm]{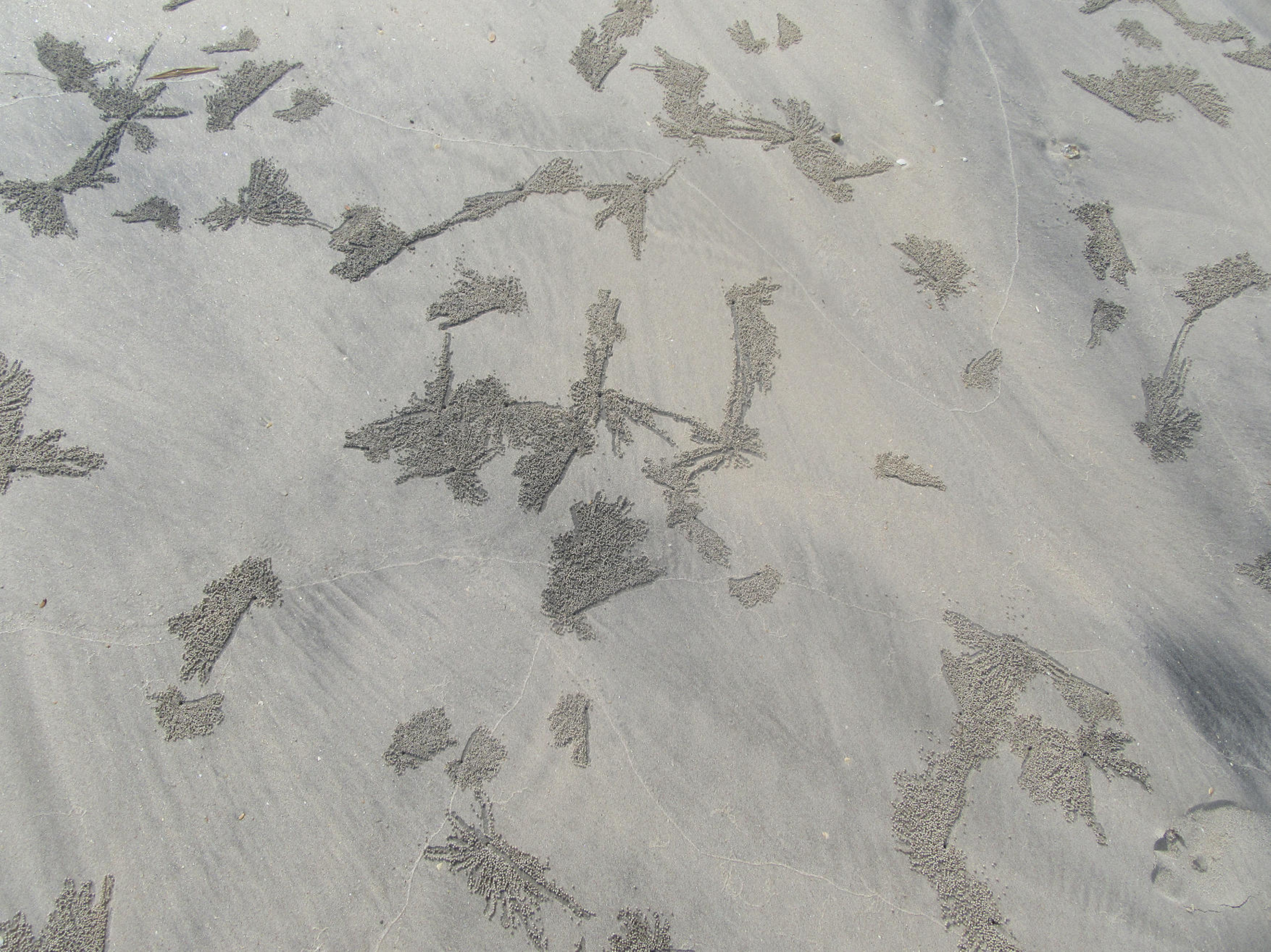}
\includegraphics[width=5.8cm, height=4.8cm]{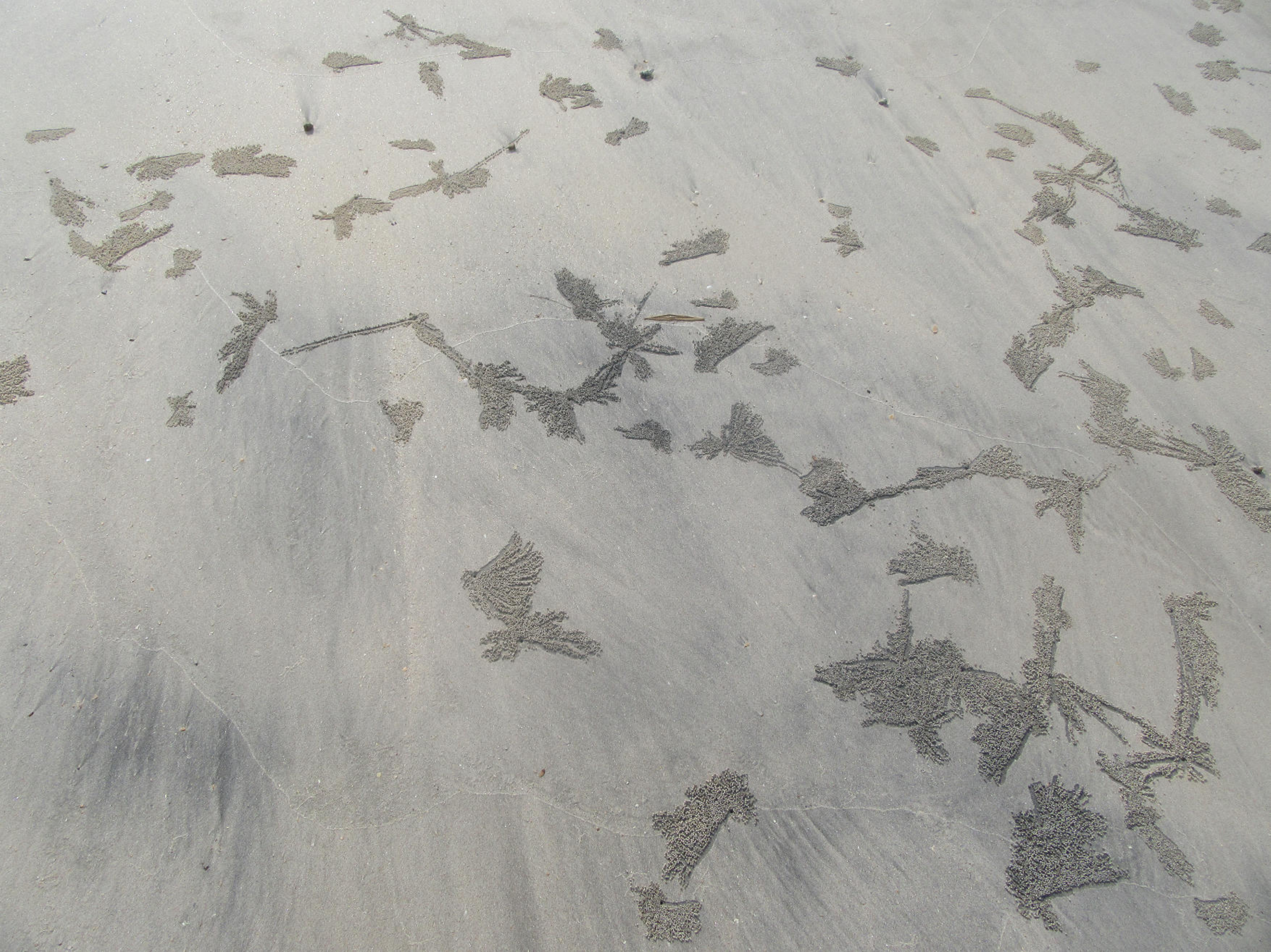}
\includegraphics[width=5.8cm, height=4.8cm]{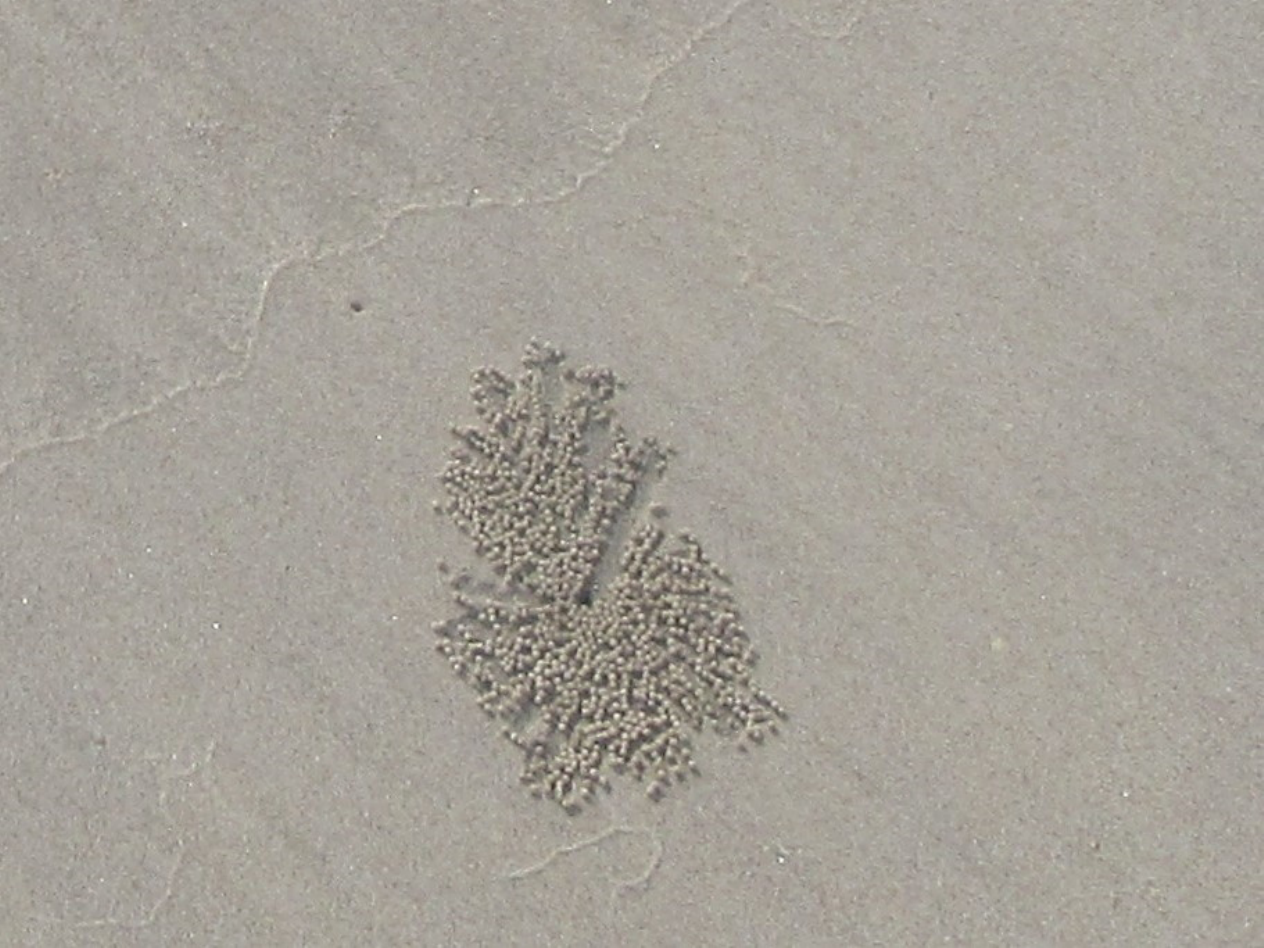}
\caption{Feeding patterns of sand--bubbler crabs, observed by the author at Tanjung Beach, Langkawi, Malaysia, in March 2017.}
\label{fig:pattern}
\end{figure}

By this process of feeding, and particularly by molding sand into pellets and distributing them, sand--bubbler crabs
create intricate patterns in the sand that 
remain there until the next high tide washes them away, see Fig. \ref{fig:pattern} that shows such feeding patterns as observed by the author at Tanjung Beach, Langkawi, Malaysia in March 2017. The patterns in the sand start from a center point, have a radial
orientation and comprise of small sand balls along a tiny path. The patterns vary in complexity depending on how many crabs there 
are contributing, on how long the last high tide is away, and on several other factors discussed below. 
The patterns in the sand are starting point and inspiration for an algorithm producing two--dimensional visual art. Before presenting such an algorithm in Sec.  \ref{sec:art}, it may be interesting to explore the feeding process and the pattern--making  a bit more deeply. 

 A main feature of the lifestyle of sand--bubblers is their burrow--orientated feeding behavior~\cite{ans88,big21,cha06,fied70}. This means that the entrance of the burrow is the center of the feeding range and the feeding process is highly structured. Basically, such a structured and stereotype  behavior can be  found generally among the  14 species of {\it Dotilla} and  17 species of {\it Scopimera} currently recognized, but there is variety  in the details of the feeding. This, in turn, offers to vary the art--making algorithm in order to achieve different graphic effects.  
 
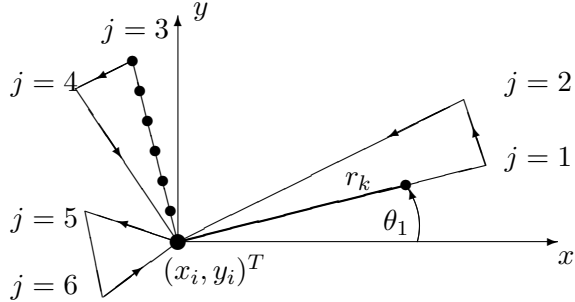
\begin{figure}[tb]
\centering
 \setlength{\unitlength}{1mm}
\begin{picture}(65,35)

\thinlines
\put(10,5){\vector(1,0){50}}
\put(10,5){\vector(0,0){30}}

\put(10,5){\line(4,1){40.5}}
\put(10,5){\line(2,1){35.8}}

\put(10,5){\line(-2,3){13.5}}
\put(10,5){\line(-1,4){6}}

\put(9,9){\circle*{1.5}}
\put(8,13){\circle*{1.5}}
\put(7,17){\circle*{1.5}}
\put(6,21){\circle*{1.5}}
\put(5,25){\circle*{1.5}}
\put(4,29){\circle*{1.5}}

\put(-0.4,20.6){\vector(2,-3){3}}

\put(10,5){\line(-4,-3){9.8}}
\put(10,5){\line(-3,1){12.2}}
\put(10,5){\vector(-3,1){8}}

\put(0.4,-2.15){\vector(4,3){5}}

\put(-2.2,9){\line(1,-5){2.28}}

\put(10,5){\circle*{2}}

\put(40,12.5){\circle*{1.5}}

\put(50.5,15.12){\line(-1,3){2.9}}
\put(50.5,15.12){\vector(-1,3){1.7}}

\put(47.6,23.82) {\vector(-2,-1){10}}

\put(4,29){\line(-2,-1){7.5}}
\put(4,29){\vector(-2,-1){4.8}}

\put(8,0){$(x_i,y_i)^T$ }

\put(53,15){$j=1$}
\put(0,32){$j=3$}
\put(-12,7){$j=5$}
\put(53,25){$j=2$}
\put(-12,-2){$j=6$}
\put(-12,25){$j=4$}

\put(60,2){$x$}
\put(12,35){$y$}

\put(32,12.5){$r_k$}
\put(37,6.5){$\theta_1$}
\put(40.58,11.77){\vector(-1,3){.1}}

\qbezier(41.5, 5)(42.5,8.75)(39.97,12.5)

\thicklines

\put(10,5){\line(4,1){30}}

\end{picture}

\caption{Algorithmically generating a pellet location with the burrow coordinates $(x_i,y_i)^T$, the trench angle $\theta_1$ of the (lower branch of the) first  trench and the radial coordinate $r_k$; see also the first term of the right--hand side of Eq. (\ref{eq:pellet}). The arrows indicate the feeding direction of a sand--bubbler crab. The outbound trench $j=3$ depicts a radial string of six equally distanced pellets, while for $j=1$ there is one pellet; for the other trenches the pellets are not shown.   }
\label{fig:burrow}
\end{figure}

The feeding starts from the burrow entrance, where the crab sets out to progress sideways along a straight line radiating from the burrow. While moving along, it sorts sand from organic--rich fine particles which are scraped off and subsequently ingested. The crab only feeds on the upper millimeter or two of the sand. The residual larger nutrient--deficient sand grains are molded  into ball--like globules, but not digested, which motivates calling them pseudo--faecal pellets. When a pellet has reached a certain diameter, which scales to the size of the crab, it is moved along the body and pushed away behind the animal. After moving along the straight line radiating from the burrow and ejecting pellets for a certain distance, the crab steps aside, turns around and moves back toward the burrow entrance while continuing with feeding and pellet ejecting. When reaching the entrance, its turns around again, and works the next line (see also Fig. \ref{fig:burrow} for a schematic description). The pellets are placed in such a way that a straight path to the burrow entrance always remains clear. This is essential for the crab to retreat to the burrow in case of external disturbance or danger. Consequently, the feeding progresses radially line by line while the lines (which are also called trenches in the biological literature) rotate around the burrow entrance just like the hands of a clock.  In fieldwork it has been observed that the rotation may be clockwise or anti--clockwise with approximately the same frequency~\cite{lus97,moh11}. By such a rotation of the feeding trench, an angular sector of a circle in the sand is excavated, cleared of eatable components, and amassed with pellets. The angular sector of uninterrupted feeding may vary; an average of $60^{\circ}$ has been reported, but also that occasionally a full circle has been completed~\cite{ans88,hart73,lus97}, see also the right--hand side of Fig. \ref{fig:pattern}. Furthermore, it has been reported that there is a fairly close relationship between the maximum length of a trench and the size of the crab. Such a relationship is plausible as  the maximum length together with the crab size scales to the maximum time needed for the crab to seek shelter in the burrow, if an external disturbance requires such an escape. For one species of sand--bubblers, {\it Dotilla wichmanni}, even a measurement of the maximum length has been recorded with the body length of eight to nine crabs~\cite{lus97}. Furthermore,   another  stereotype of pattern has been identified as sometimes the distribution of pellets along the trench follows a rhythmic pattern where the crab regularly  suspends pellet ejecting and  ``concentric rings''  emerge, as shown for {\it Dotilla clepsydrodactylus}~\cite{moh11}.

In view of the structured burrow--orientated feeding behavior of  sand--bubbler crabs as described above,   it seems not surprising that regular patterns in the sand emerge. Ideally, a well--ordered collection of full circular areas with the burrow entrance in the center should  appear that are filled with radial strings of equally distanced pellets. However, looking at patterns from nature (see, for instance, Fig. \ref{fig:pattern})  reveals them as irregular in several respects. There are reasons for these irregularities. A first reason is that for a colony of sand--bubblers the coordinates of the burrow entrances are randomly distributed across the beach where they are dwelling.  There is almost  always a certain minimum distance between two burrow entrances and the number of burrows for a given square of beach follows a Poisson distribution, as shown for  the species {\it Dotilla fenestrata}~\cite{gh01}. 
A second reason is that the orientation of the first trench (and hence the overall orientation of the angular sector) is not different from a random distribution~\cite{lus97,moh11}. There is no indication that the orientation is related to the slope of the sand surface, or any compass direction or the sun's position.  
Another reason, for instance reported for the species {\it Scopimera inflata}~\cite{fied70}, is that a crab may move a short distance  away from the entrance of the burrow before starting to feed, which results in a gap between burrow and trench. The width of the gap appears to be randomly distributed around the crab's size. A further reason is that for a given burrow (and therefore a crab of given size) the trenches have not all the same length, even if the feeding is undisturbed~\cite{gh01}. There is a weak correlation between trench length and feeding time, with the trenches growing longer the longer the crab feeds, but generally also trench length must be regarded as to follow a random distribution.  
 Finally, a crab may be externally disturbed at any time, for instance by  intruders such as other crabs  or  human observers,   or eating enemies such as shorebirds. As sand--bubblers are extremely perceptive to movement, in such a case the crab rapidly escapes to the burrow. It may reemerge a little later and resume eating. The feeding may or may not continue along the trench last used before escape. It has been convincingly argued by field experiments that sand--bubblers (for instance  the species   {\it Dotilla wichmanni}) must possess remarkable orientation abilities that allow them  to resume the direction along which they were feeding, even in the absence of visual clues such as a partly finished trench~\cite{lus97}. Whether or not a crab uses these abilities in a given situation, however, also appears to follow a random distribution. Lastly, a crab may interrupt its feeding even if there is no evident external trigger, for instance to rest or to restore its water supply, which is done in the burrow. 

To sum it up: The exact position of each pellet follows from the structure of the burrow--oriented feeding behavior, but is subject to a substantial degree of random. If we reframe the organized and sequential placement of a sufficiently large number of pellets as a dynamic process, the pattern generation could be interpreted as a random walk substantially biased by the feeding structure.  This interplay between determinism and random may be one of the reasons why the patterns in the sand appear non--trivial and of some aesthetic value.  
Finally, we may note that 
apart from feeding, there are at least two further aspects of the sand--bubbler's lifestyle that might be interesting as inspiration for visual art,  burrow--digging and interaction with other crabs. These aspects could be addressed in future work.

\section{Design of art--making algorithms} \label{sec:art}
\subsection{Generating patterns from pellet distribution} \label{sec:gener}
The feeding behavior of sand--bubbler crabs as described in Sec.  \ref{sec:beh} is now used to design templates for algorithmically generating two--dimensional visual art. With respect to the regular and irregular aspects of the sand patterns, the art--making algorithm is also comprised of deterministic and random elements.

Describing the algorithm starts with the notion that every pellet has a location $(x_{ijk},y_{ijk})^T$ in a two--dimensional plane. The index $(ijk)$ refers to the $k$--th pellet ($k=1,2,3,\ldots,K_{ij}$) belonging to the $j$--th trench ($j=1,2,3,\ldots,J_i$) of the $i$--th burrow  ($i=1,2,3,\ldots,I$). The pellet location is algorithmically specified by
\begin{equation} \left( \begin{array}{c} x_{ijk} \\ y_{ijk }\end{array}\right)=  \left( \begin{array}{c} x_i+r_k \cdot \cos{(\theta_j)} \\ y_i+ r_k \cdot \sin{(\theta_j)} \end{array}\right) +  \left( \begin{array}{c} \mathcal{N}(\mu_{ijk},\sigma_{ijk}^2) \\  \mathcal{N}(\mu_{ijk},\sigma_{ijk}^2)  \end{array}\right) , \label{eq:pellet}\end{equation}
where $(x_i,y_i)^T$ are the coordinates of the $i$--th burrow, $\theta_j$ is the trench angle belonging to the $j$--th trench, and $r_k$ is the radial coordinate of the $k$--th pellet. For each burrow, the maximum number of pellets $K_{ij}$ remains to be specified for a given $i$ and $j$; the same applies to the maximum number of trenches $J_i$. Fig. \ref{fig:burrow} unpacks the description and shows  an example with one burrow  and six trenches (three outbound and three returning).

The first term of the right--hand side of Eq.  (\ref{eq:pellet}) can be seen as to represent the deterministic aspect of the pellet ejecting that  characterizes  the burrow--orientated feeding behavior of sand--bubblers. The second term represents the random aspect with shifting the position by a realization of a random variable normally distributed with mean $\mu_{ijk}$ and variance $\sigma_{ijk}^2$. Accordingly, every pellet location $(x_{ijk},y_{ijk})^T$  has three deterministic parameters: burrow coordinates $(x_{i},y_{i})^T$, trench angle $\theta_j$ and radial coordinate $r_k$; and
two stochastic parameters:  mean $\mu_{ijk}$ and variance $\sigma_{ijk}^2$ of the normal distribution from which the random shift in position is realized.
\begin{algorithm}[tb]
\caption{Generate patterns from pellet distribution}
\begin{algorithmic}
\STATE Set maximum number of burrows $I$ 
\FOR {i=1 to I}
\STATE Generate burrow coordinates $(x_i,y_i)^T$
\STATE Set maximum number of trenches $J_i$
\FOR{j=1 to $J_i$}
\STATE Generate trench angle $\theta_j$ 
\STATE Set maximum number of pellets $K_{ij}$
\FOR{k=1 to $K_{ij}$}
\STATE Generate pellet radial coordinate $r_k$
\STATE Set mean $\mu_{ijk}$ and variance $\sigma_{ijk}^2$ 
\STATE Calculate pellet location $(x_{ijk},y_{ijk})^T$ by Eq. (\ref{eq:pellet})
\STATE Return location and store for subsequent coloring and visualization
\ENDFOR
\ENDFOR

\ENDFOR
\end{algorithmic}

\end{algorithm}
Eq. (\ref{eq:pellet}) not only specifies the location of a pellet, it can also be used to generate patterns consisting of pellets, trenches and burrows, see Algorithm 1. Therefore, for each burrow coordinates $(x_i,y_i)^T$, one sequence of trench angles $(\theta_1,\theta_2,\ldots,\theta_{J_i})$ and $J_i$ sequences of pellet radial coordinates $(r_1,r_2,\ldots,r_{K_{ij}})$ are defined.  A major guideline for defining these parameter sequences is the intention to replicate characteristic features of the sand--bubbler pattern as discussed in Sec. \ref{sec:beh}. These patterns can be labeled according to the graphic effects they convey and subsequently taken as templates to create two--dimensional visual art works. In the numerical experiments reported in Sec.   \ref{sec:examp} the following templates are used: (i) random trench length (RTL), (ii) growing trench length (GTL), (iii) concentric rings (CCR), and (iv) burrow--to--trench gaps (BTG).   Each template varies the art--making algorithm and replicates a specific pattern found in nature. For random trench length (RTL), the maximal number of trenches $J_i$ is a random integer  uniformly distributed on $[1,J_{max}]$. The first  and the last trench angle ($\theta_1$ and $\theta_{J_i}$) are chosen as realizations of a uniform random variable distributed on $[0,2\pi]$; the remaining trench angles are calculated from the number $J_i$ to be equidistant between $\theta_1$ and $\theta_{J_i}$. Each trench has a length $\ell_j$ that is a realization of a random variable normally distributed on $[\mu_T,\sigma_T^2]$. From the length $\ell_j$,  the pellet radial coordinates are calculated to be equidistant with distance $\mathit{d}_j$.  Growing trench length (GTL) is similar to RTL (and uses the same general design parameters), but  the trench length $\ell_j$ is calculated by a normal distribution where the mean $\mu_T$ grows with trench $j$ by a growth rate $\Delta \mu_T$. Thus, the first trench has  mean $\mu_T$, while the $J_i$--th trench has mean 
$\mu_T+(J_i-1) \Delta \mu_T$. The variance $\sigma_T^2$ is constant.  Concentric rings (CCR) again have trenches with constant mean and variance but do not  always have equidistant pellet radial coordinates as there are $\lambda$ gaps of width $g_\lambda$ where no pellets are placed. Finally, burrow--to--trench gaps (BTG) start with a gap of width $g_0$ in pellet placement.    Tab. \ref{tab:design} summarizes the design parameters.
 \begin{table}[tb]
\caption{Design parameters and default values for generating patterns from pellet distribution, see Algorithm 1. The parameters of RTL are general and also apply to the other templates.} \label{tab:design}
\begin{tabular}{|l|l|l|c|}
\hline
Template & Design parameter & Symbol & Default value \\ \hline \hline
RTL & Number of burrows & $I$ & $3$ \\
(and & Burrow coordinates & $(x_i,y_i)^T$ &  $\mathcal{N}(0,7)$\\
general)& Maximum number of trenches& $J_{max}$& $50$\\
 & Pellet distance& $d_j$ & $0.25$ \\
& Mean random shifting & $\mu_{ijk}$ & $-1, 0, 1$\\
& Variance random shifting & $\sigma_{ijk}^2$ & $0.3, 0.8$\\
& Mean trench length &$\mu_T$& $25$\\
& Variance trench length & $\sigma_T^2$ & $1$\\
\hline
GTL & Grow rate & $\Delta \mu_T$ & $2$\\
\hline
CCR &Number of gaps& $\lambda$& $3$\\
&Gap width& $g_\lambda$& $4$\\
\hline
BTG &Burrow gap width & $g_0$& $8$\\
\hline

\end{tabular}

\end{table}

Algorithm 1 specifies the location of the pellets and hence the structure of the patterns. In nature, the patterns are monochromatic as the pellets all have  the color of the sand they are built from, see Fig. \ref{fig:pattern}. The artistic interpretation of the patterns suggests using different colors for making them more visually appealing and also conveying additional information. As the algorithm imposes a chronological order on pellet placement by a forward counting of burrows, trenches and finally the pellets themselves, a possible design is to use a colorization that reflects this order. Alternatively, 
each burrow can be associated with a specific color and the order of the pellet placement can be characterized by shades or brightness of this color.  In addition, the burrows using the same template (see Tab. \ref{tab:design}) may take colors that are similar.

\subsection{Analysis by computational aesthetic measures} \label{sec:measure}
Next, the templates and their design parameters are analyzed by computational aesthetic measures. This analysis aims at studying two questions. (i) To what extend do the measures vary over the templates and parameters? This goes along with assessing how  sensitive these relationships are.  (ii) Which templates and/or parameters are most suitable to guide and control the art--making process toward desired values of the  aesthetic measures?   

Three computational aesthetic measures are studied, Benford's law measure (BFL),  Ross, Ralph and Zong's bell curve measure (RRZ) and  a fractal dimension measure (FRD), which were chosen  because a recent study reported them to be weakly correlated~\cite{den14}. All measures are calculated for the images with $512 \times 512$ pixels.  The Benford's law measure $\text{BFL}$~\cite{den14,neu17}  is known for scaling to the naturalness of an image. It is calculated by taking the distribution of  the luminosity of an image and comparing it to the Benford's law distribution:
\begin{equation} \text{BFL}= 1- d_{total}/d_{max}\end{equation}
where  $d_{total}= \sum_{i=1}^9 \left(H_{I}(i)-H_{B}(i) \right)$,  $d_{max}=1.398$ is the maximum possible value of $d_{total}$, $H_B=\left( 0.301,0.176,0.125,0.097,0.079,0.067,0.058,0.051,0.046 \right)$  is the Benford's law distribution  and
 $H_{I}$ is the normalized sorted $9$--bin histogram of the luminosity of the image. The luminosity  of the image is calculated for each pixel of the image by taking the weighted sum of the red ($R$), green ($G$) and blue ($B$) values: 
$\text{lum}=0.2126 \cdot R+0.7152 \cdot G+0.0722\cdot B$. 

Ross, Ralph and Zong's bell--curve measure $\text{RRZ}$~\cite{den14,ross06} is a measure of color gradient normality. It compares the color gradients of the image to a normal (bell-curve or Gaussian) distribution. Therefore, the color gradient of the colors red, green and blue are calculated for each pixel, summarized and normalized by a detection threshold, see \cite{den14,ross06} for details. Hence, we obtain a distribution with mean $\mu$ and variance $\sigma^2$. The distribution is discretized into $100$ bins to obtain a discrete  color gradient distribution $p_i$.  For $\mu$ and $\sigma^2$ calculated from the observed color gradient distribution, an expected discrete Gaussian distribution $q_i$ is computed.  Finally, the  Kullback--Leibler divergence between these two distributions gives the bell--curve measures 
\begin{equation} \text{RRZ}= \sum_i p_i \log{\left( p_i/q_i \right)}. \label{eq:kld} \end{equation}
Note that in deviation from~\cite{den14,ross06},  the Kullback--Leibler divergence in Eq. (\ref{eq:kld}) is not amplified by the factor 1000.  

The third  aesthetic measure studied is the fractal dimension $\text{FRD}$~\cite{den14,spe03} of the image.  Following empirical studies and an argument by Spehar et al.~\cite{spe03}, a fractal image of dimension $1.3 \leq d \leq 1.5$ is most preferred by human evaluation, compared to images that have a fractal dimension outside this range.  Thus, a fractal dimension $d \approx 1.35$ is considered to be most desirable and the fractal aesthetic measure 
\begin{equation} \text{FRD} = \max{\left(0,1-| 1.35-d | \right)} \end{equation}
can be defined, 
where $d$ is the fractal  dimension of the image calculated by box--counting. 
\begin{figure}[tb]
\includegraphics[trim = 20mm 95mm 50mm 90mm,clip,width=5.8cm, height=4.8cm]{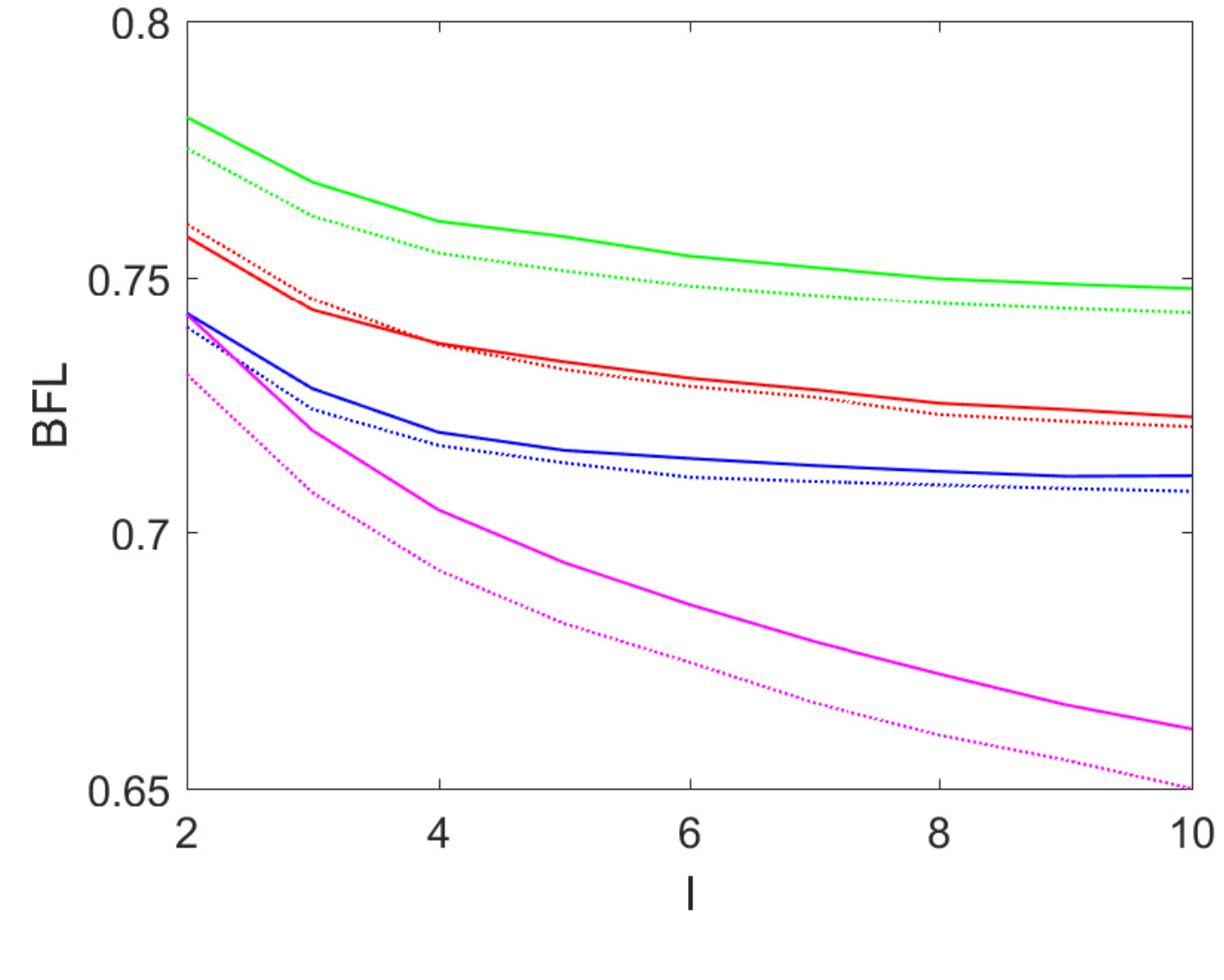}
\includegraphics[trim = 20mm 95mm 50mm 90mm,clip,width=5.8cm, height=4.8cm]{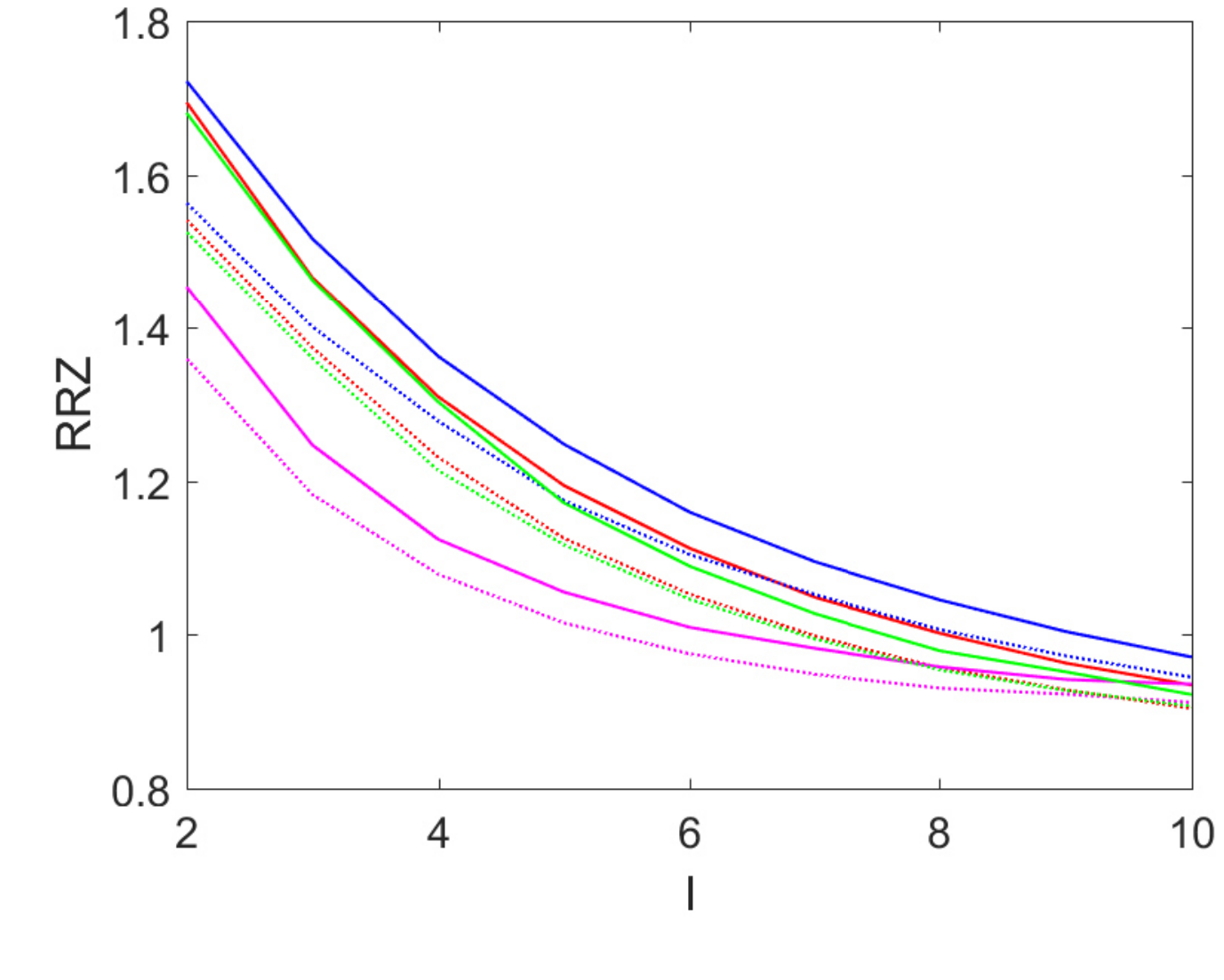}
\includegraphics[trim = 20mm 95mm 50mm 90mm,clip,width=5.8cm, height=4.8cm]{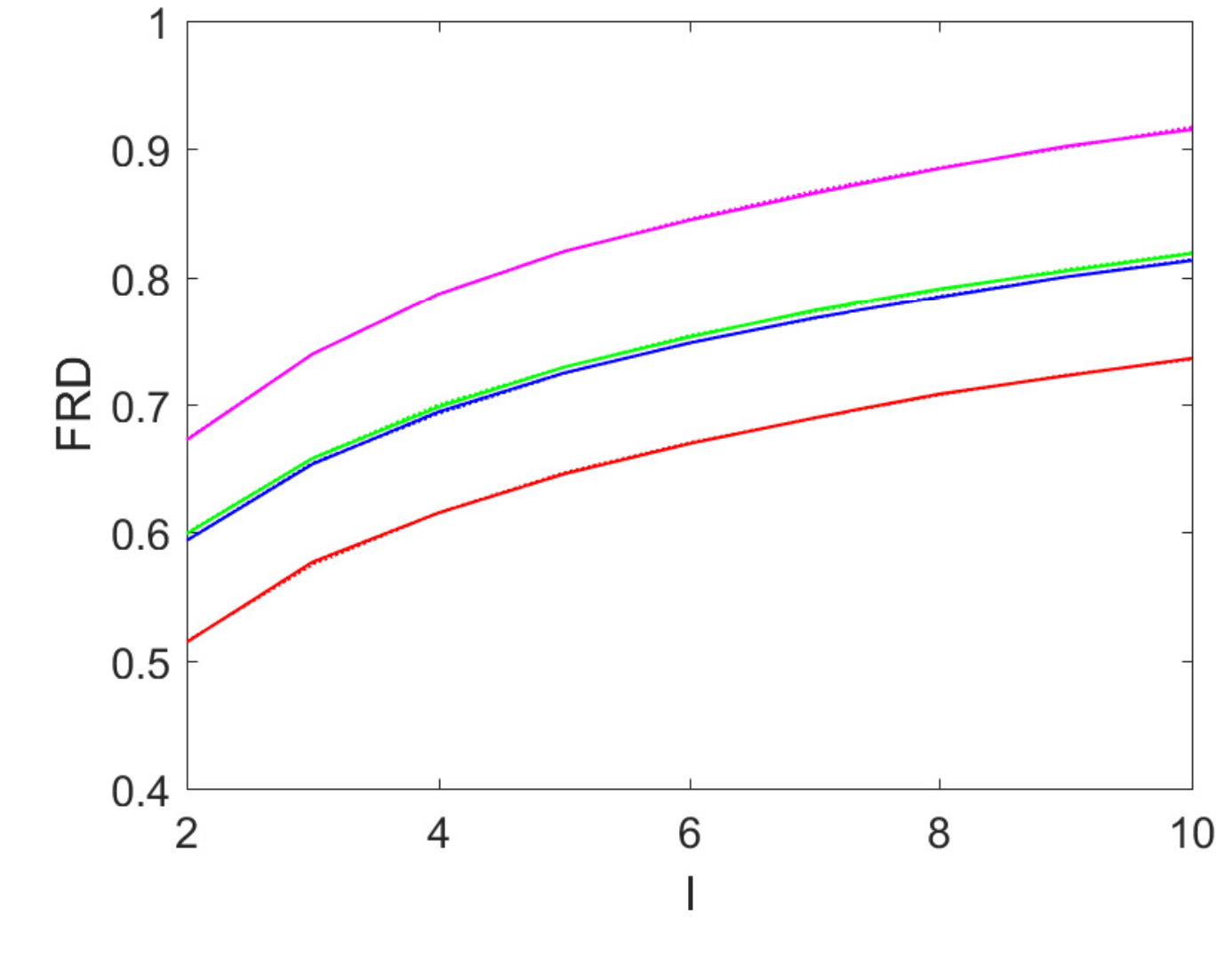}

\includegraphics[trim = 20mm 95mm 50mm 90mm,clip,width=5.8cm, height=4.8cm]{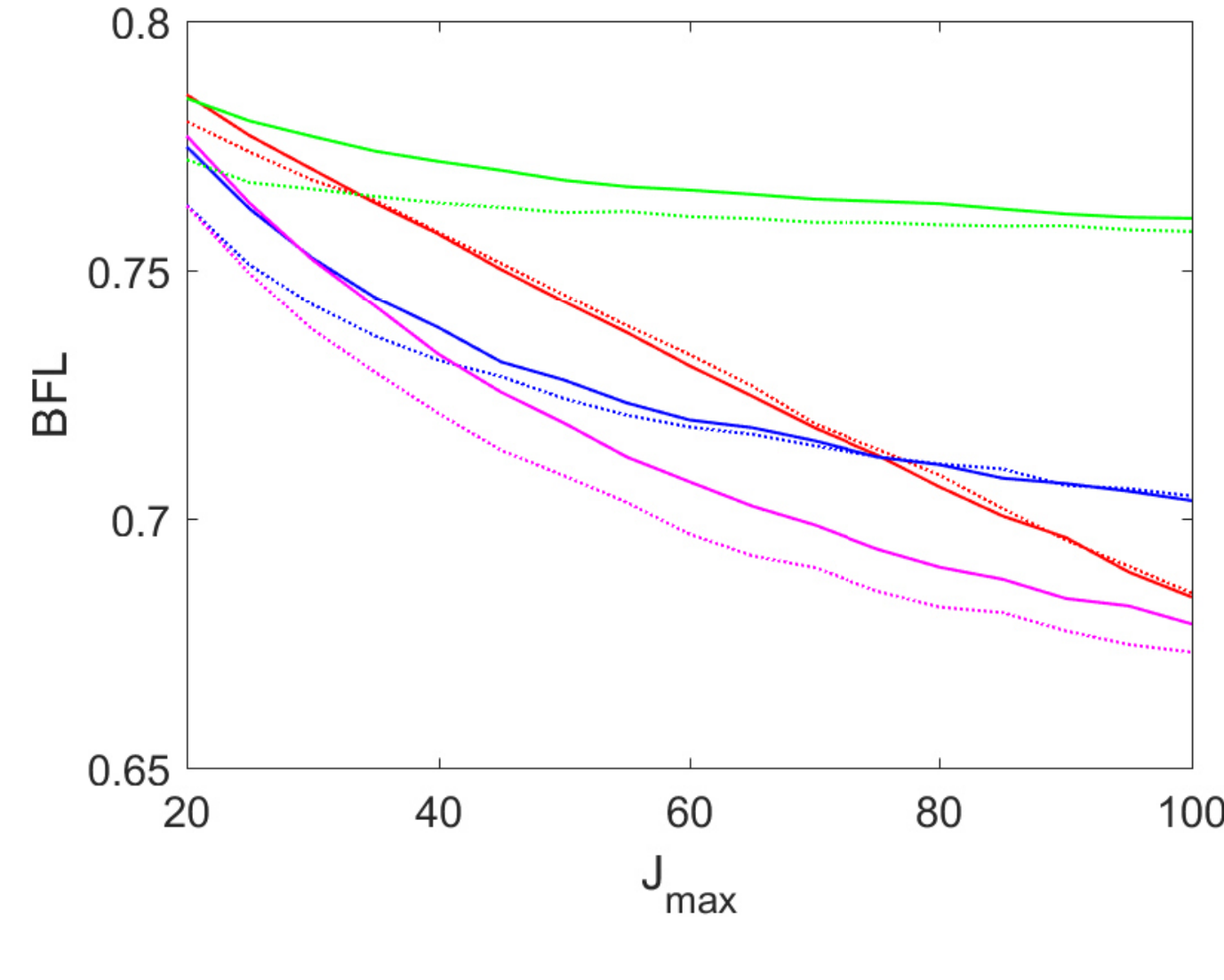}
\includegraphics[trim = 20mm 95mm 50mm 90mm,clip,width=5.8cm, height=4.8cm]{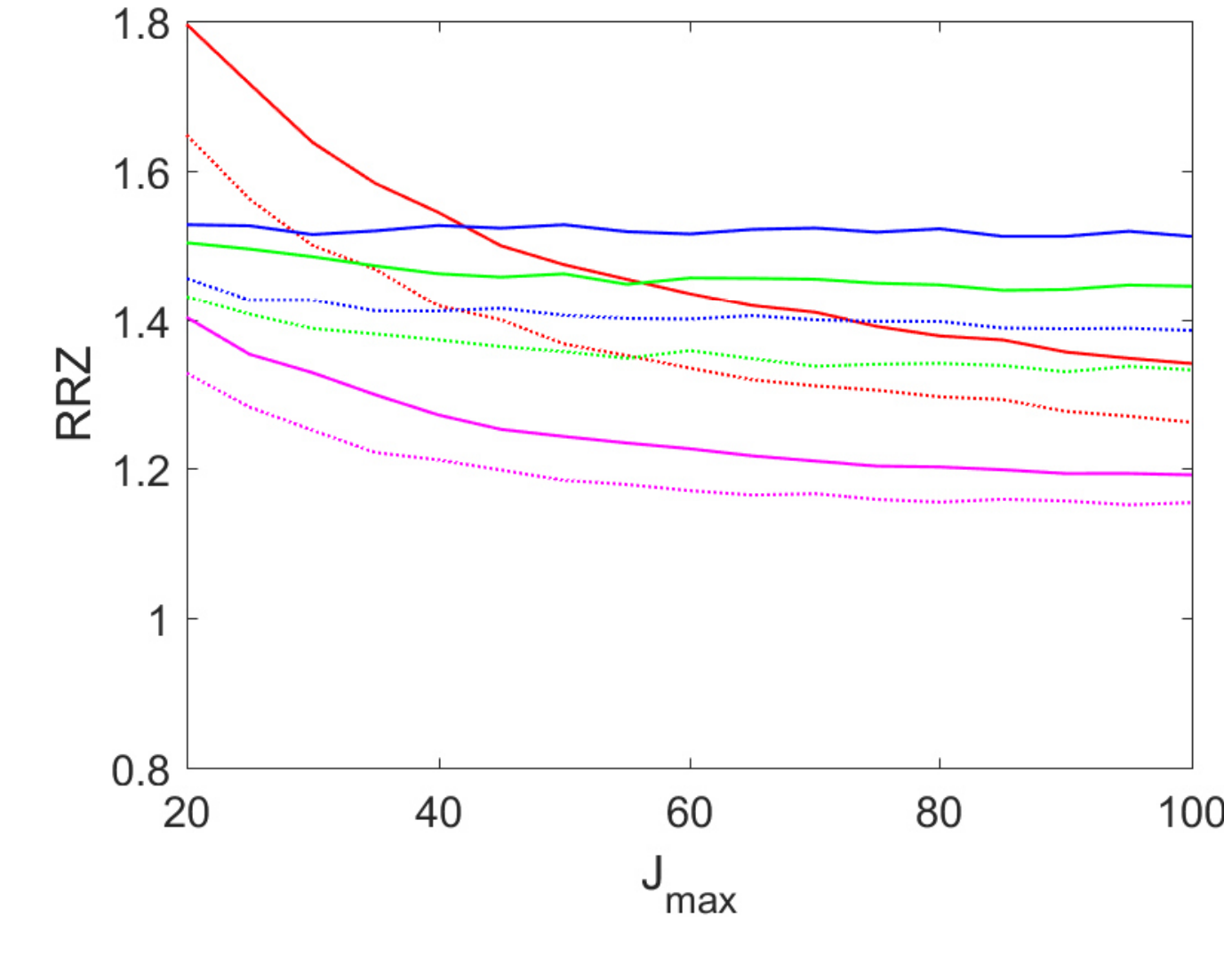}
\includegraphics[trim = 20mm 95mm 50mm 90mm,clip,width=5.8cm, height=4.8cm]{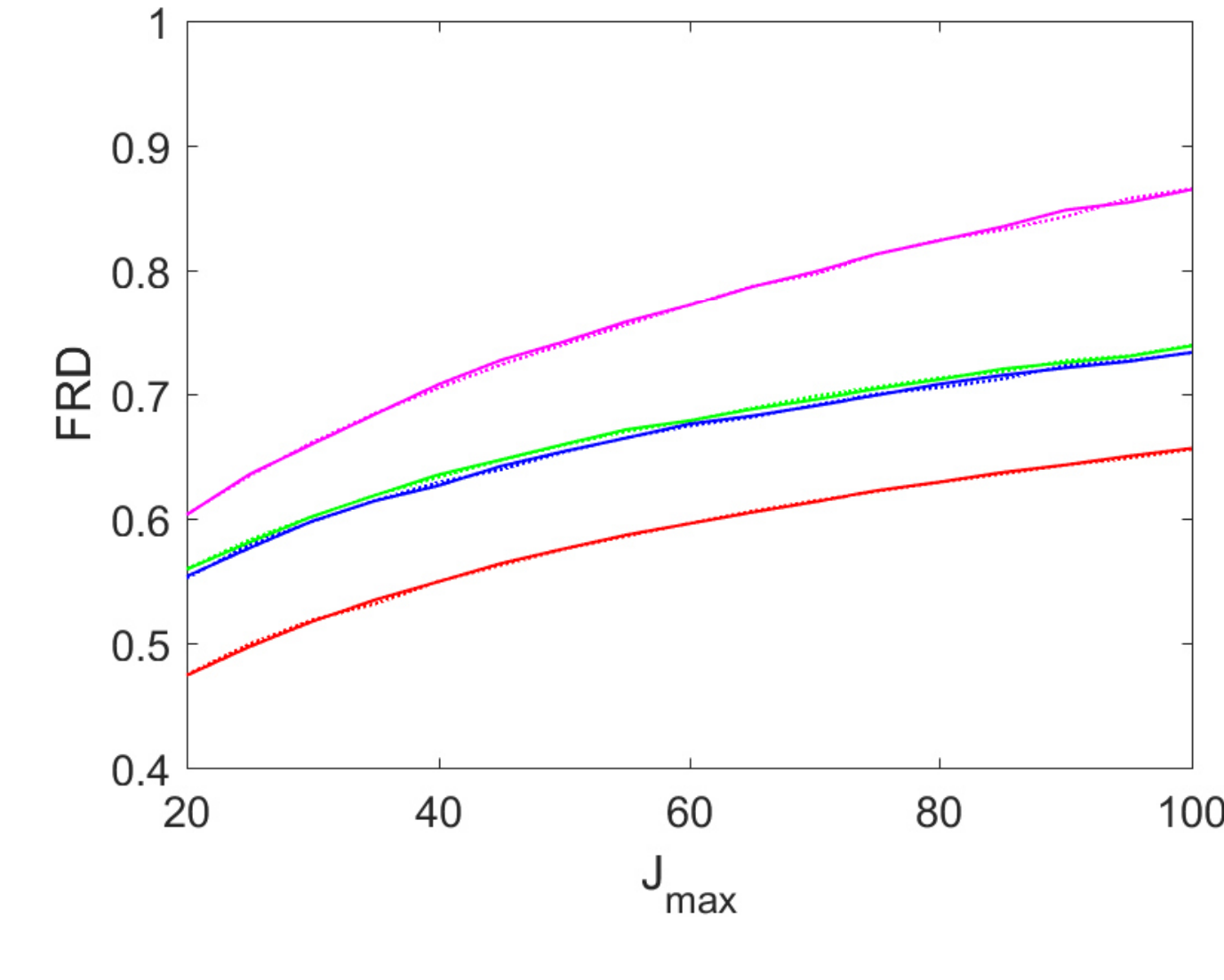}

\includegraphics[trim = 20mm 95mm 50mm 90mm,clip,width=5.8cm, height=4.8cm]{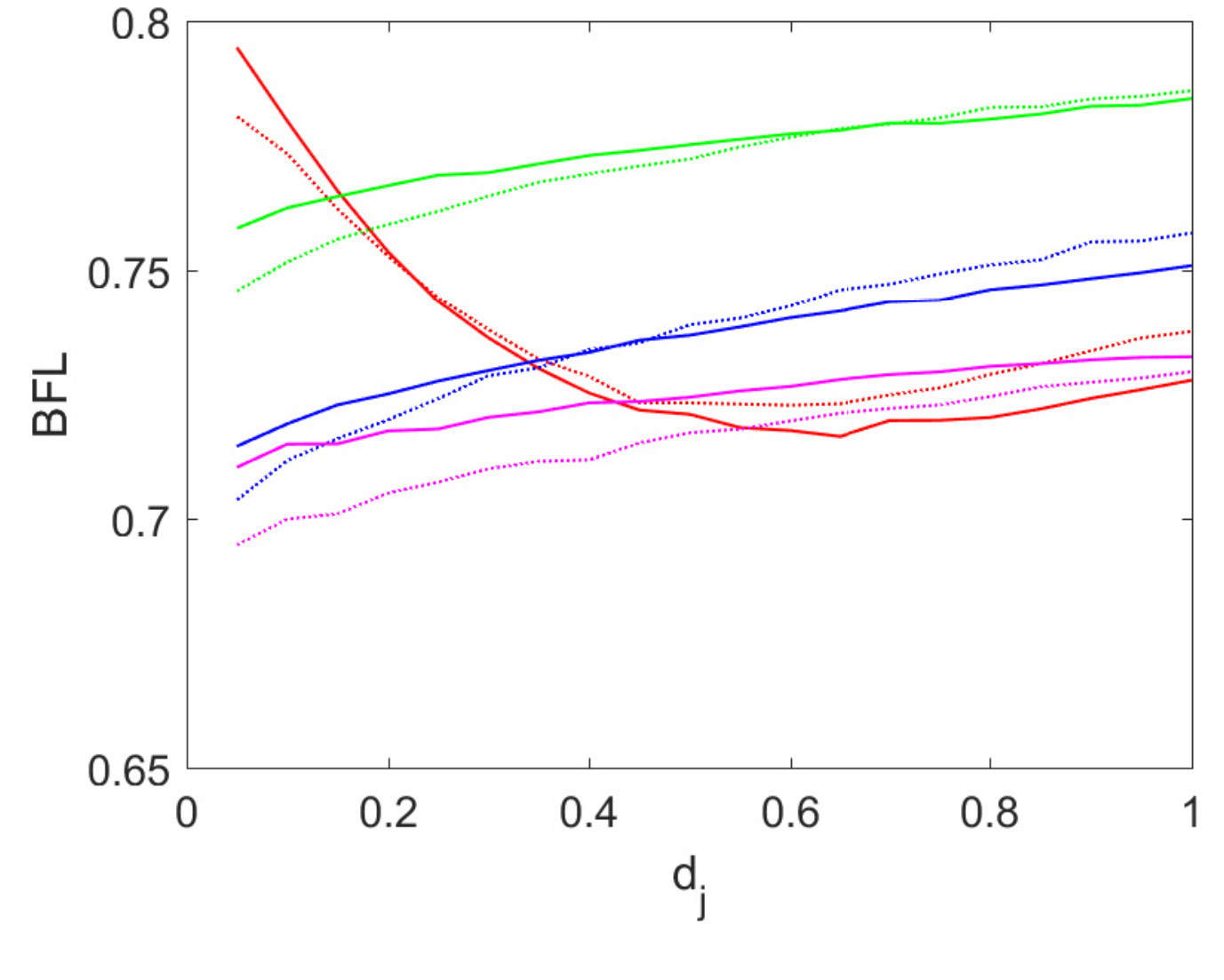}
\includegraphics[trim = 20mm 95mm 50mm 90mm,clip,width=5.8cm, height=4.8cm]{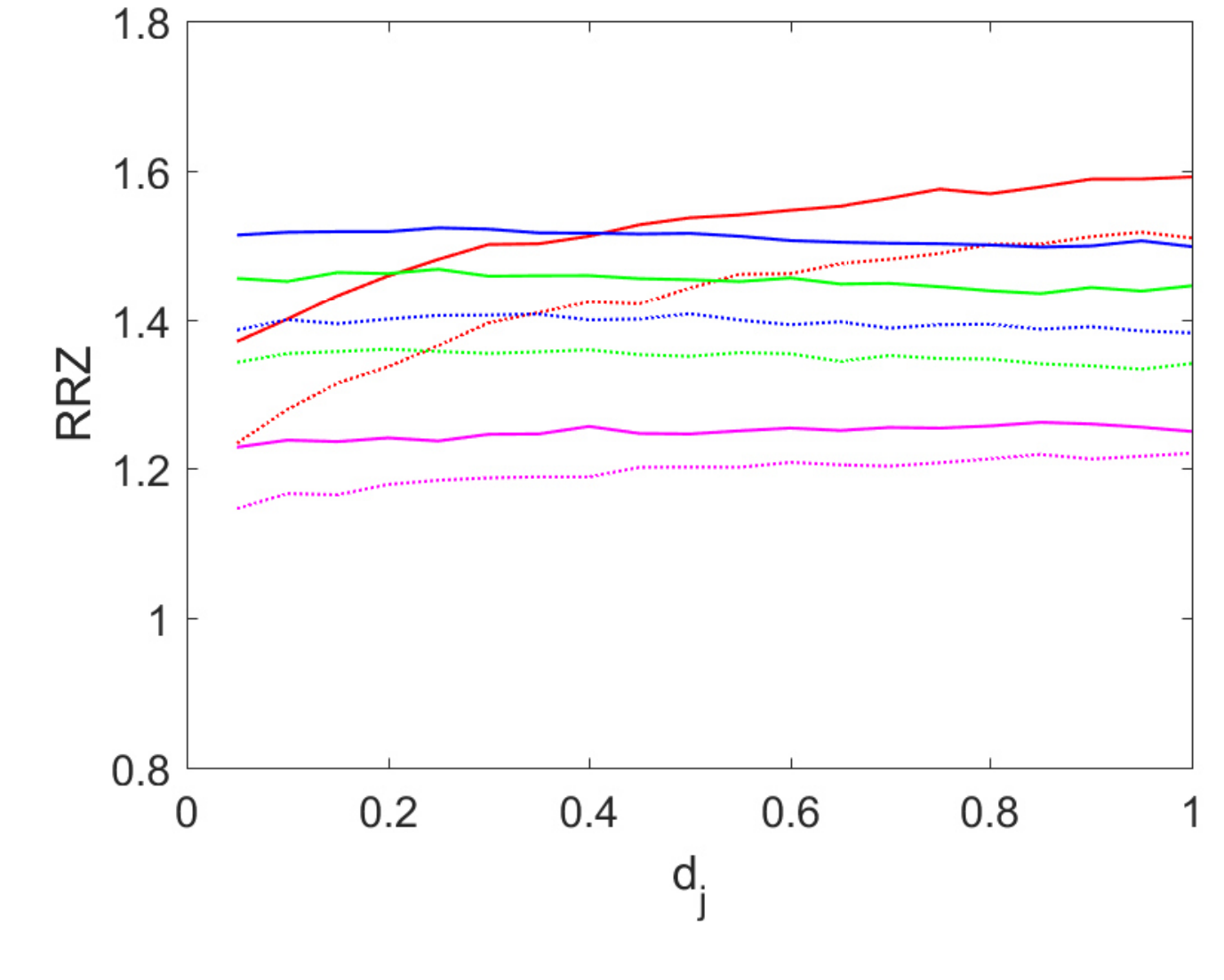}
\includegraphics[trim = 20mm 95mm 50mm 90mm,clip,width=5.8cm, height=4.8cm]{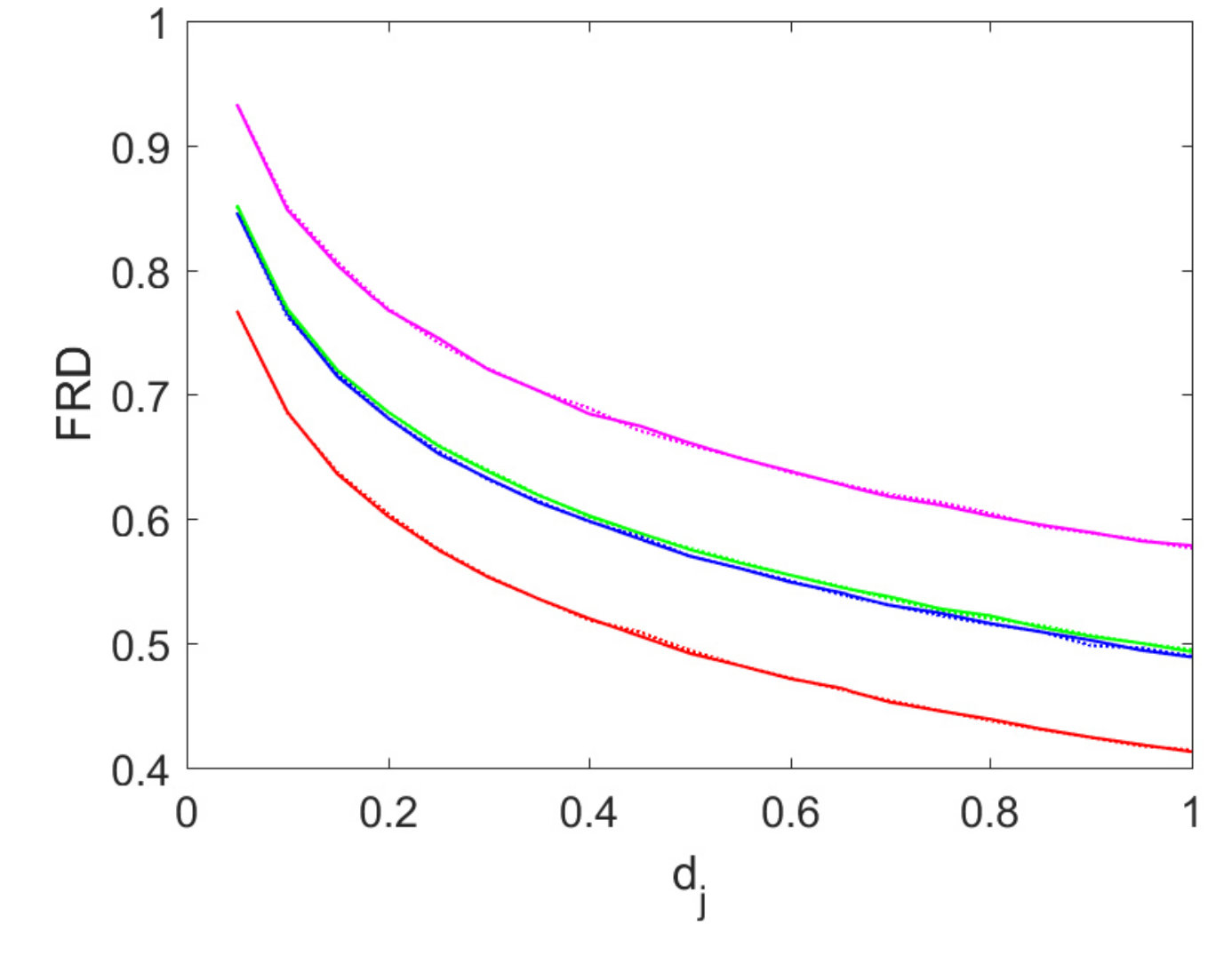}

\caption{Aesthetic measures as a function of design parameters:  number of burrows $I$,  maximum number of trenches $J_{max}$ and pellet distance $d_j$. 
RTL (blue),  GTL (magenta), CCR (red),    BTG (green);  solid lines, $\sigma_{ijk}^2=0.3$, dotted  lines, $\sigma_{ijk}^2=0.8$.  
}
\label{fig:measures}
\end{figure}

Fig. \ref{fig:measures} shows the measures for the templates RTL, GTL, CCR and BTG for selected design parameters and two values of the variance of the random shifting  $\sigma_{ijk}^2$.  The results are means over $100$ images generated by Algorithm 1 and the design parameters listed in Tab. \ref{tab:design}. The design parameter number of burrows $I$ is varied over $2 \leq I \leq 10$, the maximum number of trenches $J_{max}$ over $20 \leq J_{max} \leq 100$ and the pellet distance $d_j$ over $0.05 \leq d_j \leq 1$.
The default values of these parameters all lie within the  range studied. Furthermore, as the aesthetic measures are calculated from the RGB values of the pixels, the colors of the image have a profound effect on the values of the measures. As discussed in Sec. \ref{sec:gener}, the colorization of the pellets is also a subject of algorithmic design. Thus,  to counter the effect of a given colorization and for not singularizing particular colors, the results are calculated as means over $70$ hue values  randomly selected. 

From the results in Fig.  \ref{fig:measures}, we see that the different templates (RTL, GTL, CCR, BTG) generally produce characteristic curves over the design parameters for the aesthetic measures BFL and RRZ. Also, the two values of random shifting (which can be interpreted as noise levels) are clearly   distinct. The general trend is that BFL and RRZ decrease with increasing number of burrows $I$ and increasing maximum number of trenches $J_{max}$, while there is an increase with increasing pellet distance $d_j$. An interesting exception is the template CCR which for varying pellet distance $d_j$ behaves differently and crosses the curves of the other templates. The results for the aesthetic measures FRD are slightly different. For FRD, the templates RTL and BTG (blue and green lines) have almost the same results and for all templates, different noise levels have only negligible effect.  A possible explanation is that the difference between RTL and BTG is only that in the images there is a circle free from pellets around the burrow entrance for BTG. The overall structure of the pellet placement is the same, which is recognized by the calculation of the fractal dimension accounting mainly for density of geometrical objects but  less for their layout. The same may apply for varying the noise level.
Based on these results assessing how sensitive the relationships between design parameters and aesthetic measures are, some conclusions can be drawn about possible ways to guide and control the art--making algorithm. Basically, there are two options. One is to switch to a particular template but keep the values of design parameters, another is to keep the template but change the values of  design parameters. (In fact, it would also be thinkable to change both template and parameter value at the same time, but this may appear rather contrived.)  For instance, to increase the aesthetic measure BFL, employing the template BTG is suitable, while RRZ is largest for RTL (but not very clearly) and FRD for GTL. Another possibility to raise BFL is to decrease the maximum number of trenches $J_{max}$. The same is also useful to increase RRZ, particularly for the template CCR, but not for FRD. There are more subtle schemes to manipulate the measures, but it is also clear that the three aesthetic measures studied here  react differently, yet even contradictory to these manipulations.  With this in mind, next section discusses possible designs of such manipulations and their effects on the art--making process.

\subsection{Guiding and controlling generative evolution} 
Main features of generative and evolving digital art are that the creating process develops over time,  is guided and controlled by the algorithmic design, and thus gains functional autonomy to produce complete art works~\cite{gal16}. A frequently used design for guidance and control is to implement a feedback, for instance via  
 evaluating the works by using a measure of their aesthetic value~\cite{mach14,rom08}. Such ideas are, for instance, implemented by employing evolutionary algorithms for finding ``optimal'' values of the aesthetic measures, which has shown to create interesting works of art~\cite{ali17,den14,hous15,rom08}. 
Algorithm 1 given in Sec. \ref{sec:gener} does not really, in itself, evoke such a perspective of autonomously guiding and controlling generative evolution. The algorithm is suitable for producing visual art works. It also depends on selecting templates and design parameters (see Tab. \ref{tab:design}), which enables modifying the works by changing the selection.   Thus, promoting  generative evolution essentially requires measuring the aesthetic value of the algorithmically generated patterns and designing a feedback.

 Of course, it is possible to evaluate the pattern  by  interaction with humans. However, there are some issues with interactive evaluation~\cite{gal10,gal12,rom08}. The first is that  humans evaluating art are usually much slower in doing so than computers algorithmically generating the works, which creates the proverbial ``fitness bottleneck''. A second reason is  that human evaluation may change over time, for instance caused by fatigue, but also by boredom, which may bias the results towards superficial novelty rather than overall quality.  A third reason is that human evaluation of art is always conditioned by personal and cultural ``taste''.  For this reason, it is no real help to distribute the evaluation to a larger group of human evaluators. In such a case,  either the tastes of different subgroups become visible (as for instance shown for evaluating the beauty of abstract paintings~\cite{mal14}), or generally the averaging effect of asking a large group of people about their aesthetic choices is reproduced, but arguably   ``the unique kind of vision expected from artists''~\cite{gal10} is not achieved.
\begin{algorithm}[tb]
\caption{Guide and control  pattern--making}
\begin{algorithmic}
\STATE Set up look--up table with template and design parameters that increase computational aesthetic measures
\STATE Calculate expected aesthetic measures for the look--up table
\STATE Select random template
and take default design parameter
\WHILE {Termination criterion not met}
\STATE Generate burrow by Algorithm 1
\STATE Calculate aesthetic measures
\IF {Measure $<$ Expectation}
\STATE Select randomly increasing template or design parameter
\ELSE
\STATE Keep template and parameter
\ENDIF
\ENDWHILE
\end{algorithmic}
\end{algorithm}

An alternative are computational aesthetic measures~\cite{den14,gal10,gal12,mal14,neu17}, which enable an automated rating without human interference or supervision. Sec. \ref{sec:measure} discussed such measures for evaluating sand--bubbler patterns. These aesthetic measures of two--dimensional visual art are frequently calculated by analyzing the (spatial) distribution of colored pixels (or groups of pixels). In other words, computational aesthetic measures use properties such as pixel--wise color information to deduce an overall rating of an image. However, there is a significant conceptual gap between measuring computational image properties and aesthetic beauty.  Apart from the measures considered in Sec.  \ref{sec:measure}, there is a huge number of measures that try to bridge this gap, see e.g.~\cite{gal12,den14,john16,mal14} for an overview. Some of them are strongly correlated, while others address different aspects of the color distribution.  Moreover, recent studies~\cite{den14,mal14} suggested that it seems rather difficult to establish stable correlations between  beauty ratings by humans and image properties, which casts some doubts on whether an evaluation of the aesthetic value of visual art based on  computational  aesthetic measures is really objective, meaningful and feasible. These studies have also shown that
computational aesthetic measures account more for certain visual effects (or ``visual styles''~\cite{den14}) than for aesthetic value in general. For instance,  Benford's law measure (BFL) assigns high values for images that have a grainy texture, while Ross, Ralph and Zong's bell curve measure (RRZ) likes distinct color progression and the fractal dimension measure (FRD) values low colorfulness highly; see den Heijer \&  Eiben~\cite{den14} for a discussion and comparison of 7 computational aesthetic measures. 
\begin{figure}[tb]
\includegraphics[trim = 20mm 95mm 50mm 90mm,clip,width=5.8cm, height=4.8cm]{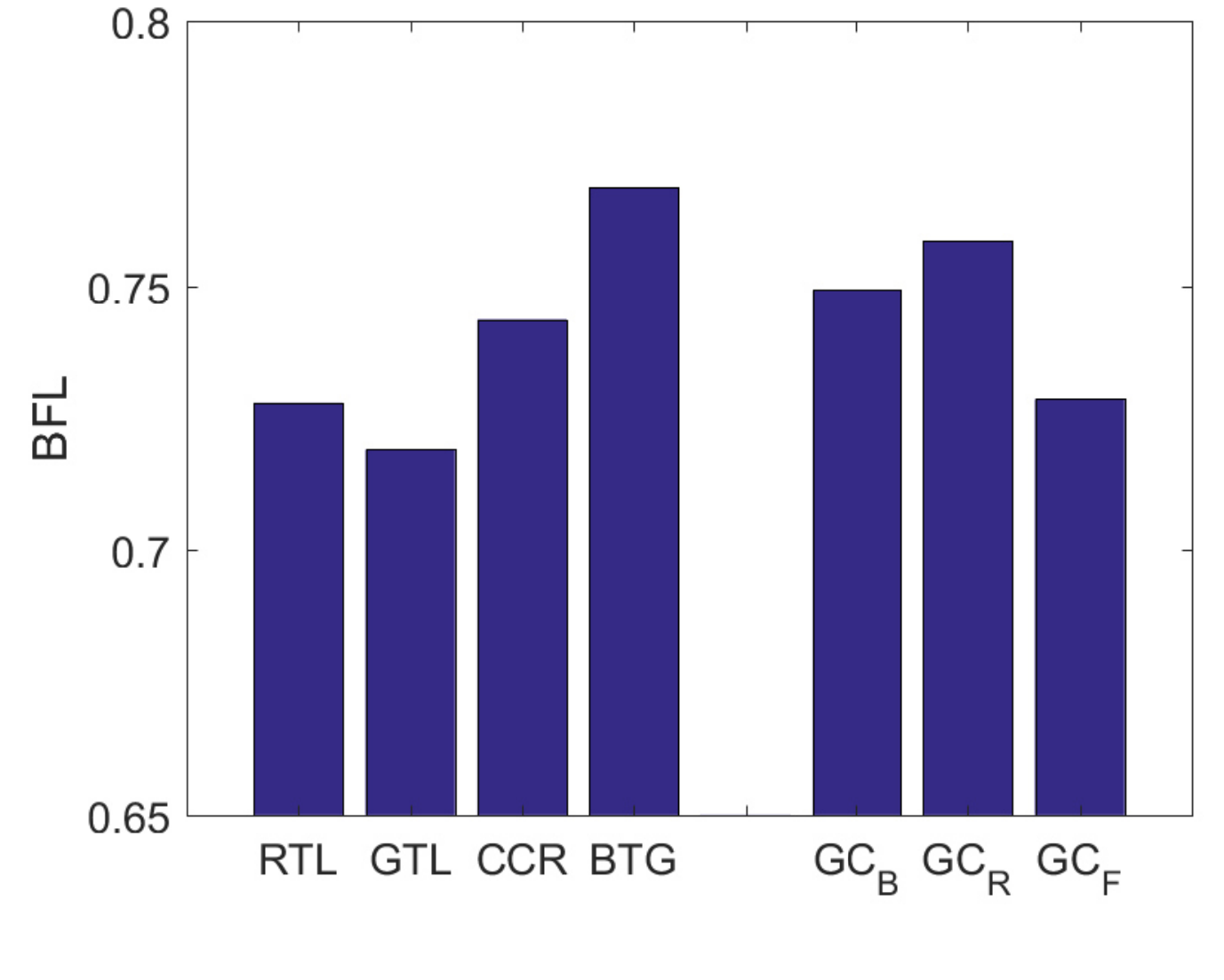}
\includegraphics[trim = 20mm 95mm 50mm 90mm,clip,width=5.8cm, height=4.8cm]{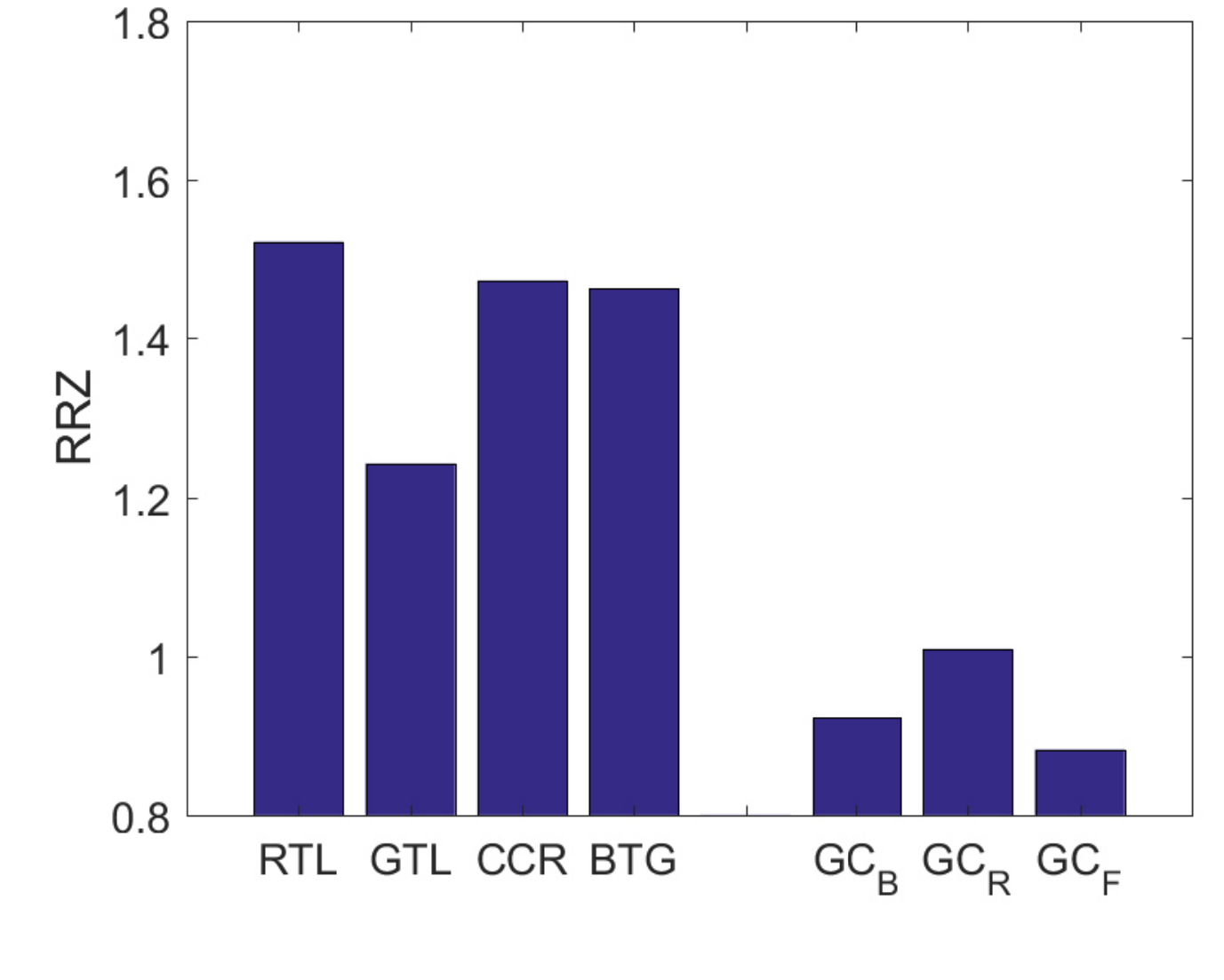}
\includegraphics[trim = 20mm 95mm 50mm 90mm,clip,width=5.8cm, height=4.8cm]{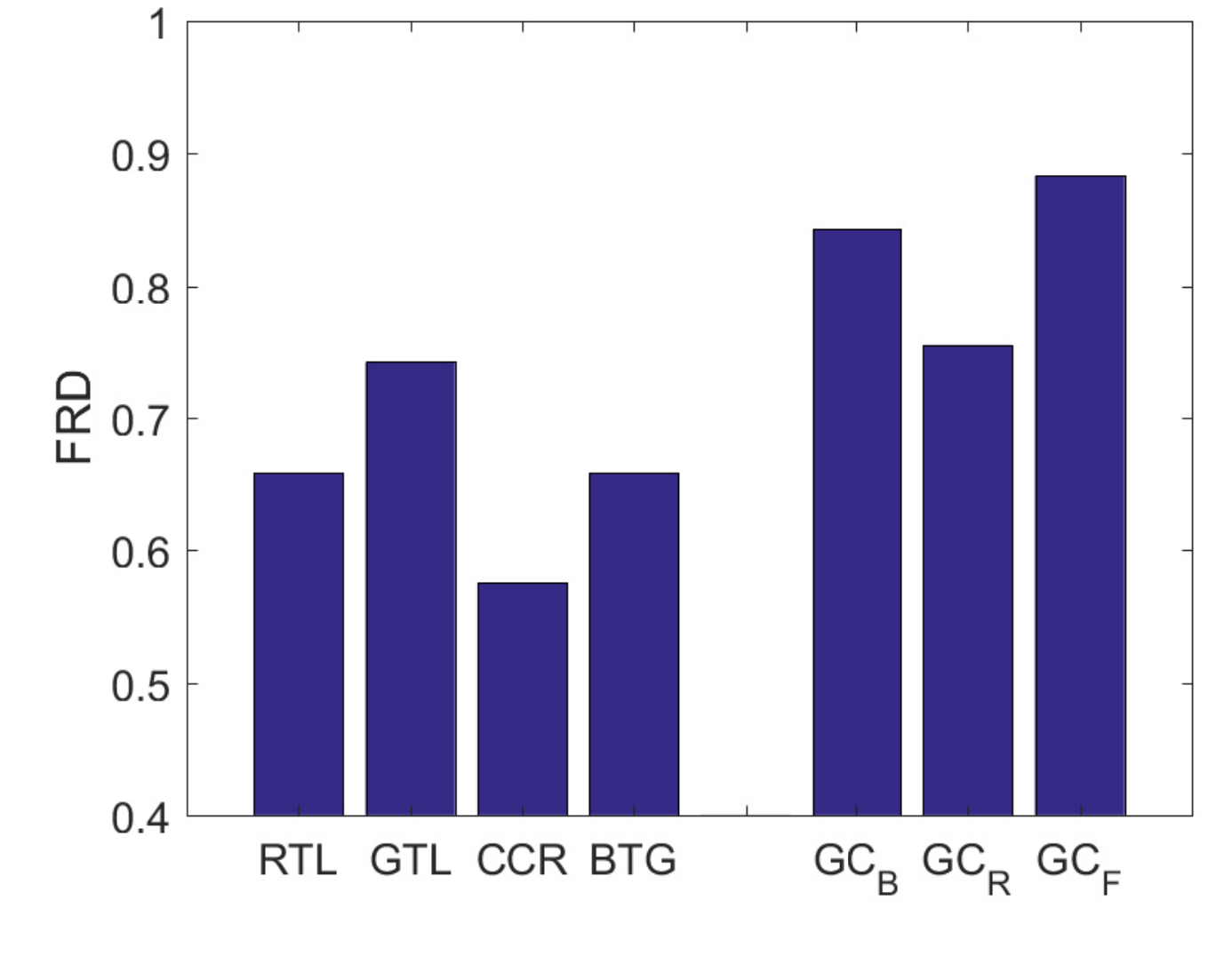}

\caption{Aesthetic measures for 
guiding and controlling pattern--making.  The measures for the templates (RTL, GLT, CCR, BTG) with the default values of the design parameters (see Tab. \ref{tab:design}) are compared to the measures obtained by Algorithm 2. GC\textsubscript{B} means using the measure BFL to guide and control, GC\textsubscript{R} uses RRZ, GC\textsubscript{F} uses FRD.  
}
\label{fig:measures1}
\end{figure}

In the following, we study an alternative approach for guiding and controlling generative evolution that tries to address the desire to promote certain visual effects, see Algorithm 2. It utilizes the relationships between design parameters and aesthetic measures discussed in Sec.  \ref{sec:measure}.    Therefore, a look--up--table is set up that lists templates and design parameters that increase aesthetic measures. After generating the pattern for each single burrow, an aesthetic measure is calculated and compared to an ``expected value'' recorded before, see Fig. \ref{fig:measures}. If the value is below  the expectation, a measure--increasing template or design parameter is activated. If not, the generation process continues unaltered.  Which template or parameter is selected is due to chance. These steps are repeated until a termination criterion is met. In this implementation, the pattern generation ends if a maximum number of burrows are placed in the image. (In the rare case that no measure--increasing template or parameter is available, the setting is also kept.)
An interesting feature of such an algorithmic design is that it immediately allows to control and intervene while the image is being created. In other words, the control does not wait until the  image is completed, but uses feedback for intervening during the process creating the art works. 
Fig. \ref{fig:measures1} shows the aesthetic measures BFL, RRZ and FRD for Algorithm 2.  The experimental setup is the same as for the templates, see Fig. \ref{fig:measures}. Again, the results are averages over 100 images and 70 random hue values.  The aesthetic measures for the templates in Tab. \ref{tab:design} are compared to using 
 the measures BFL (GC\textsubscript{B}), RRZ (GC\textsubscript{R}) and FRD (GC\textsubscript{F}) for guiding and controlling the art--making process. Apart from the results of the measures  that are obtained by using the same measures to guide and control, we also record the results of the other measures to examine cross--effects (GC\textsubscript{R} and GC\textsubscript{F} for BFL, GC\textsubscript{B} and GC\textsubscript{F} for RRZ and  GC\textsubscript{B} and GC\textsubscript{R} for FRD). 
 The results in Fig. \ref{fig:measures1} indicate that using templates and design parameters is mostly suitable to manipulate the aesthetic measures in an intended way. For BFL we obtain that the measure for GC\textsubscript{B} is increased as compared to the templates RTL, GTL and CCR, while the images do contain a mix of all templates. For FRD, we even find an increase in GC\textsubscript{F} compared to all templates. However, for RRZ we get a rather strong decrease in GC\textsubscript{R}. A possible explanation is that the measure RRZ   accounts for   color gradient normality, while the effect of Algorithm 2 is basically in changing the selection of templates and design parameters but only indirectly in colorization. Hence, the measure RRZ is poorly sensitive to such a changed selection.     These speculations are supported by looking at cross--effects as we see that also the measures that are not used for guidance and control show similar results. This additionally allows to conjecture that for images such as the feeding patterns  the aesthetic measures are more correlated than initially assumed.     Next, we look at the visual results produced by both algorithms\footnotemark\footnotetext{See https://feit-msr.htwk-leipzig.de/sandbubblerart/ for further images and videos.}.

\subsection{Examples of art works} \label{sec:examp}
\begin{figure}[tb]
\includegraphics[trim = 40mm 95mm 40mm 100mm,clip,width=5.8cm, height=4.8cm]{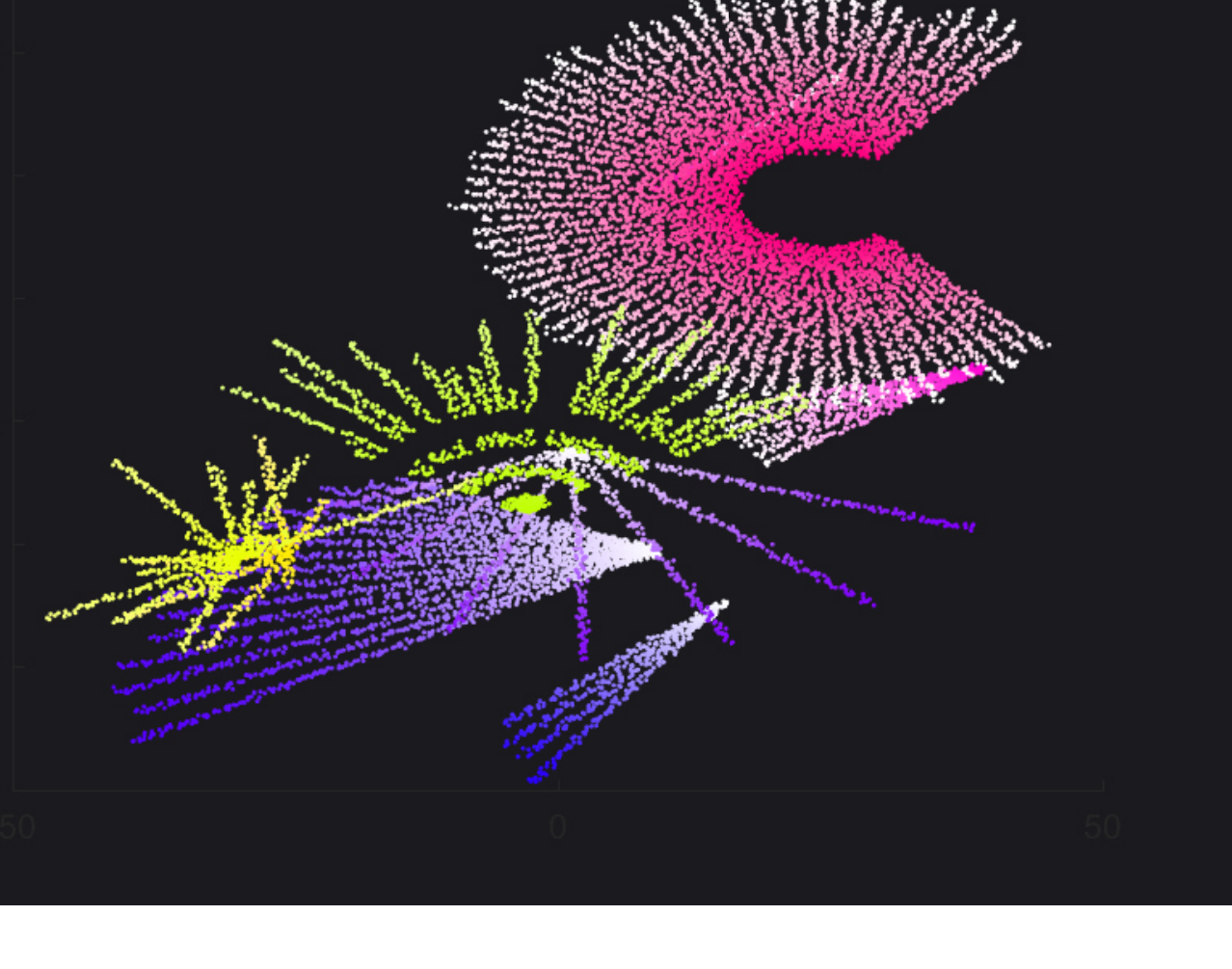}
\includegraphics[trim = 40mm 95mm 40mm 100mm,clip,width=5.8cm, height=4.8cm]{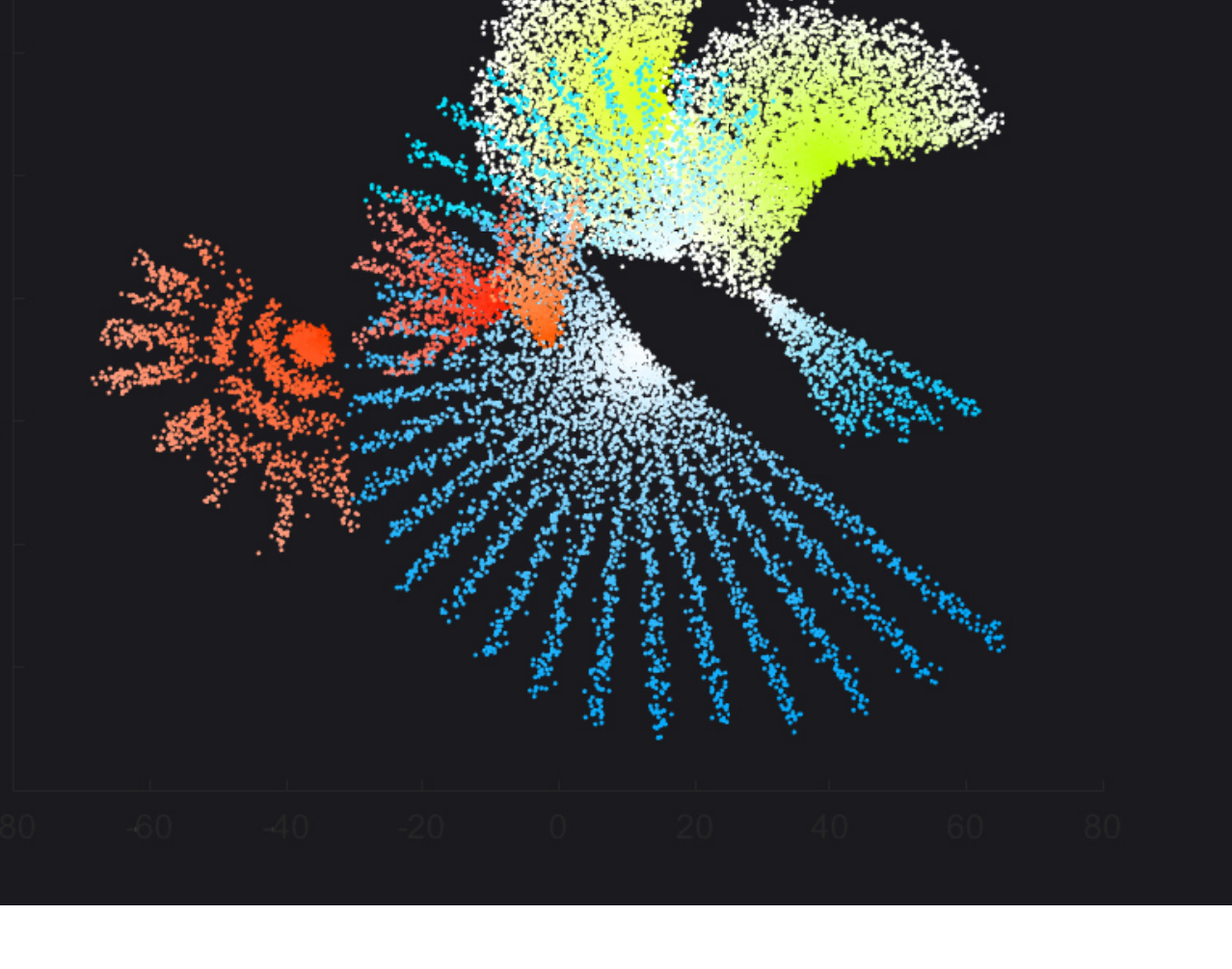}
\includegraphics[trim = 40mm 95mm 40mm 100mm,clip,width=5.8cm, height=4.8cm]{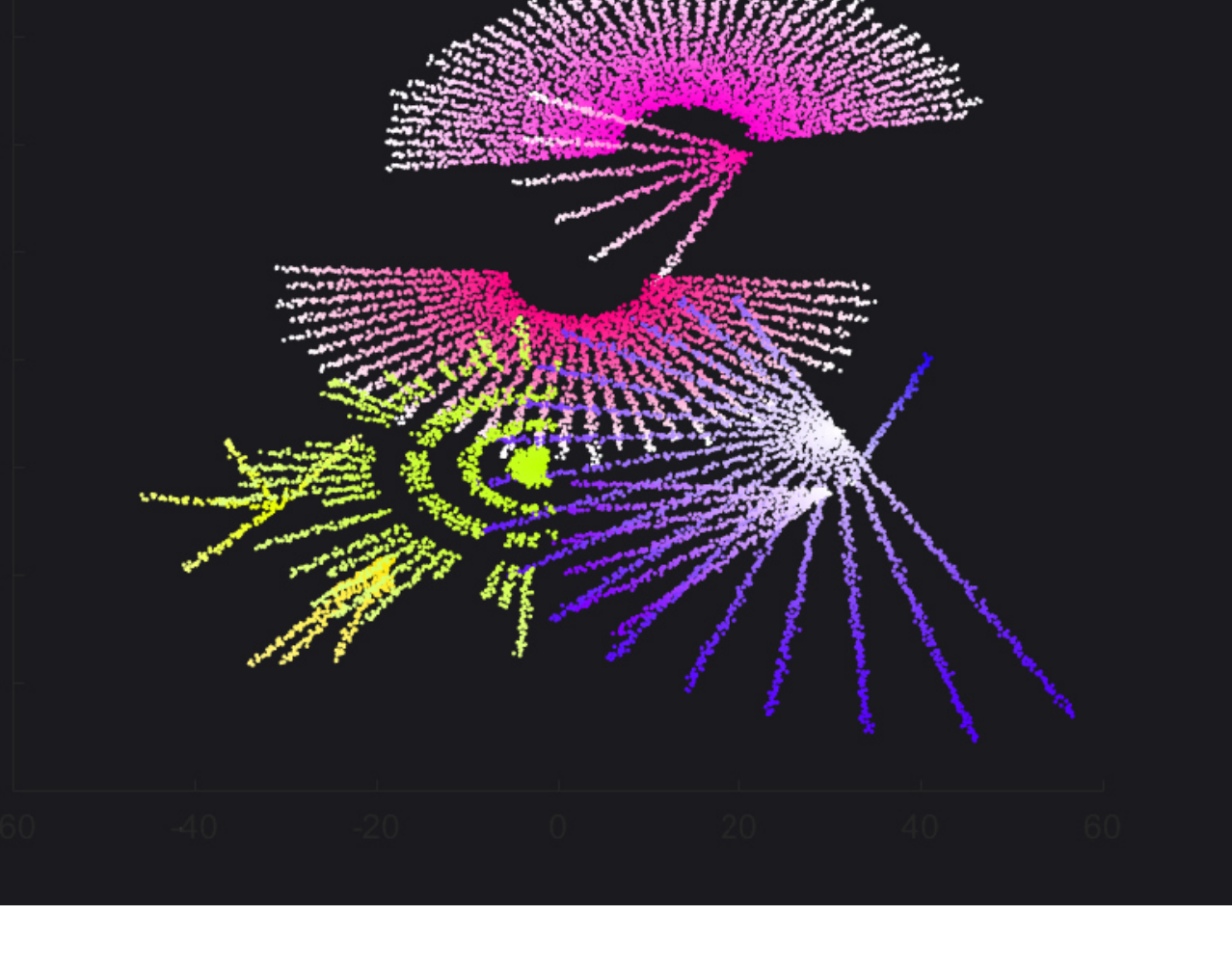}

\includegraphics[trim = 40mm 95mm 40mm 100mm,clip,width=5.8cm, height=4.8cm]{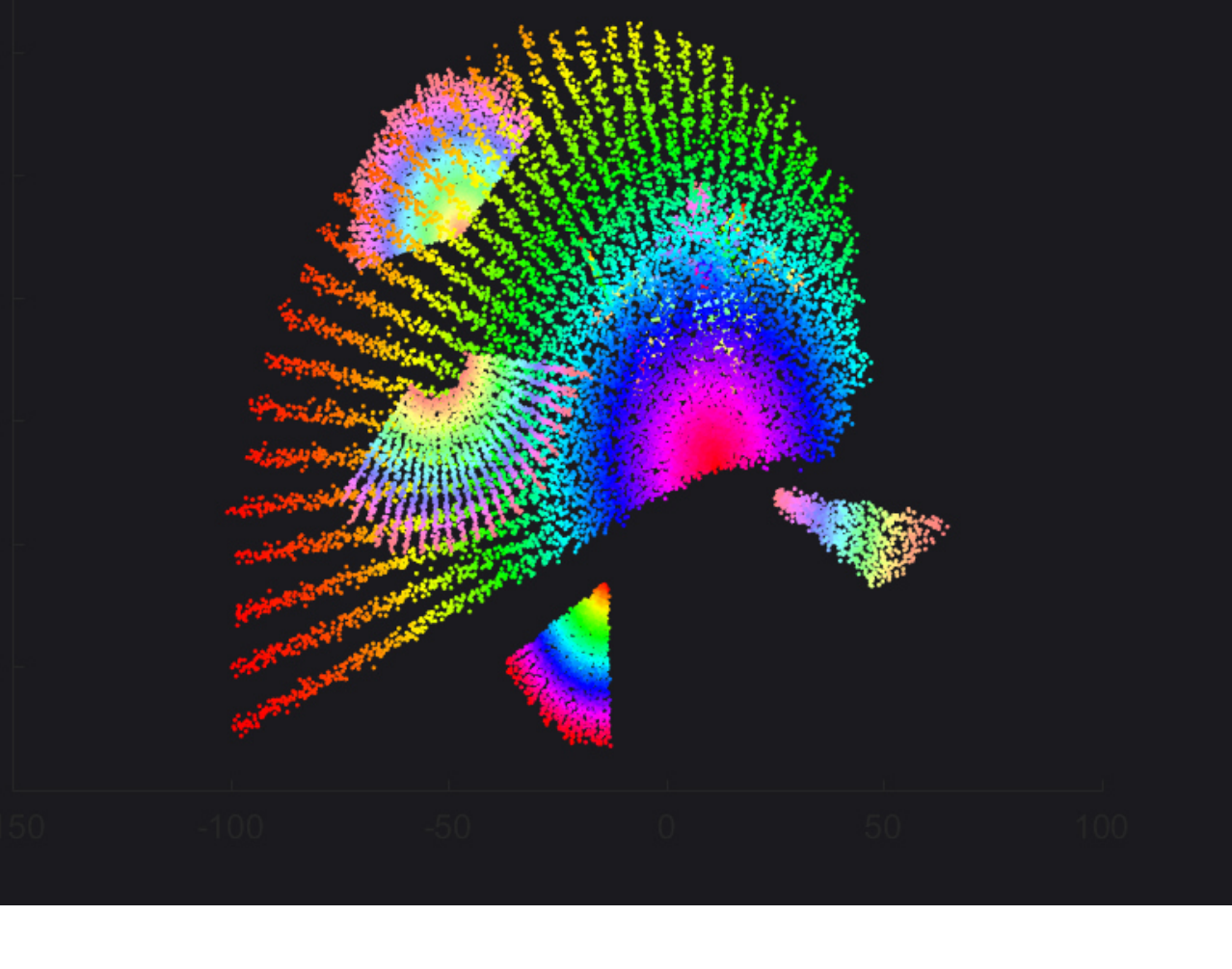}
\includegraphics[trim = 40mm 95mm 40mm 100mm,clip,width=5.8cm, height=4.8cm]{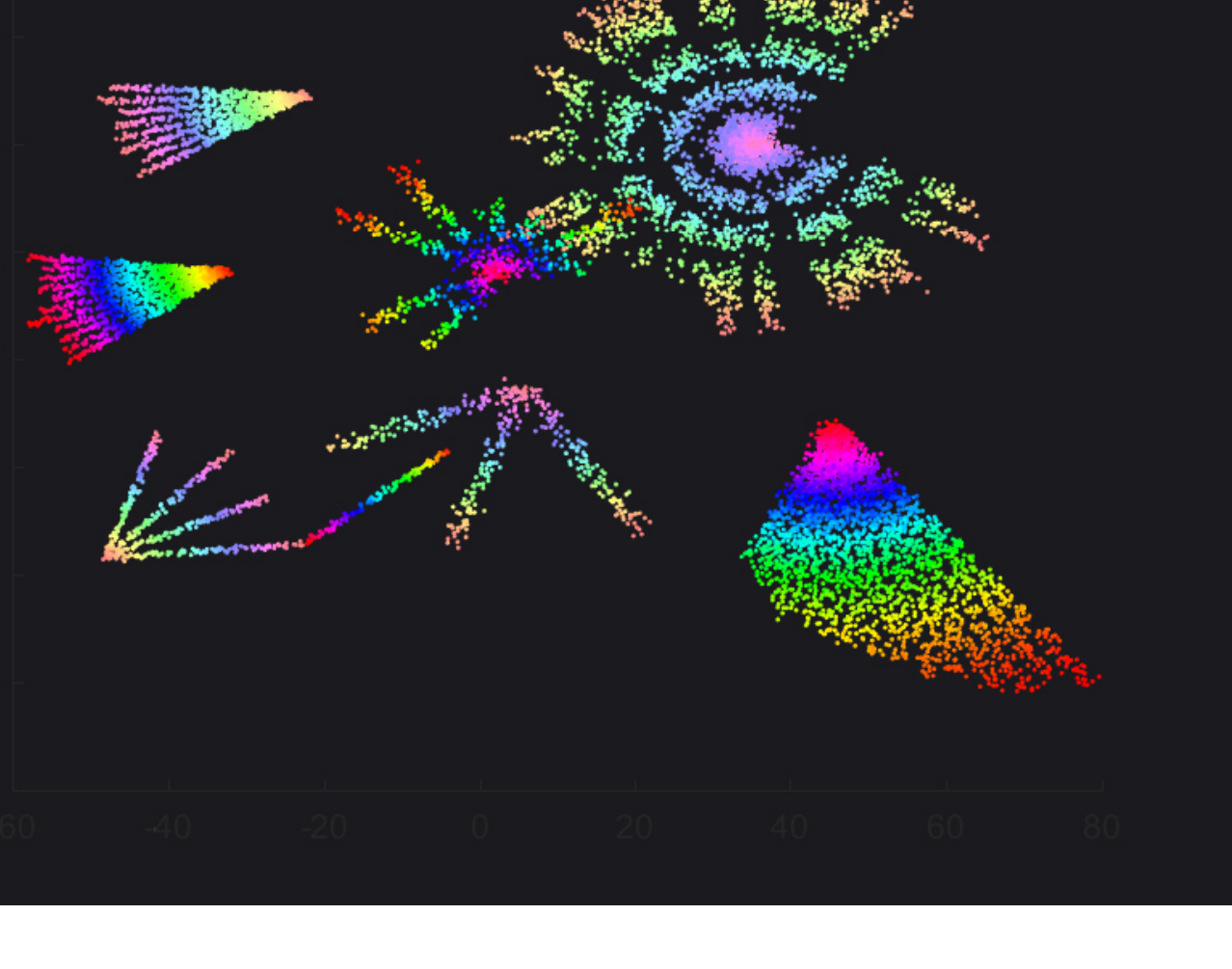}
\includegraphics[trim = 40mm 95mm 40mm 100mm,clip,width=5.8cm, height=4.8cm]{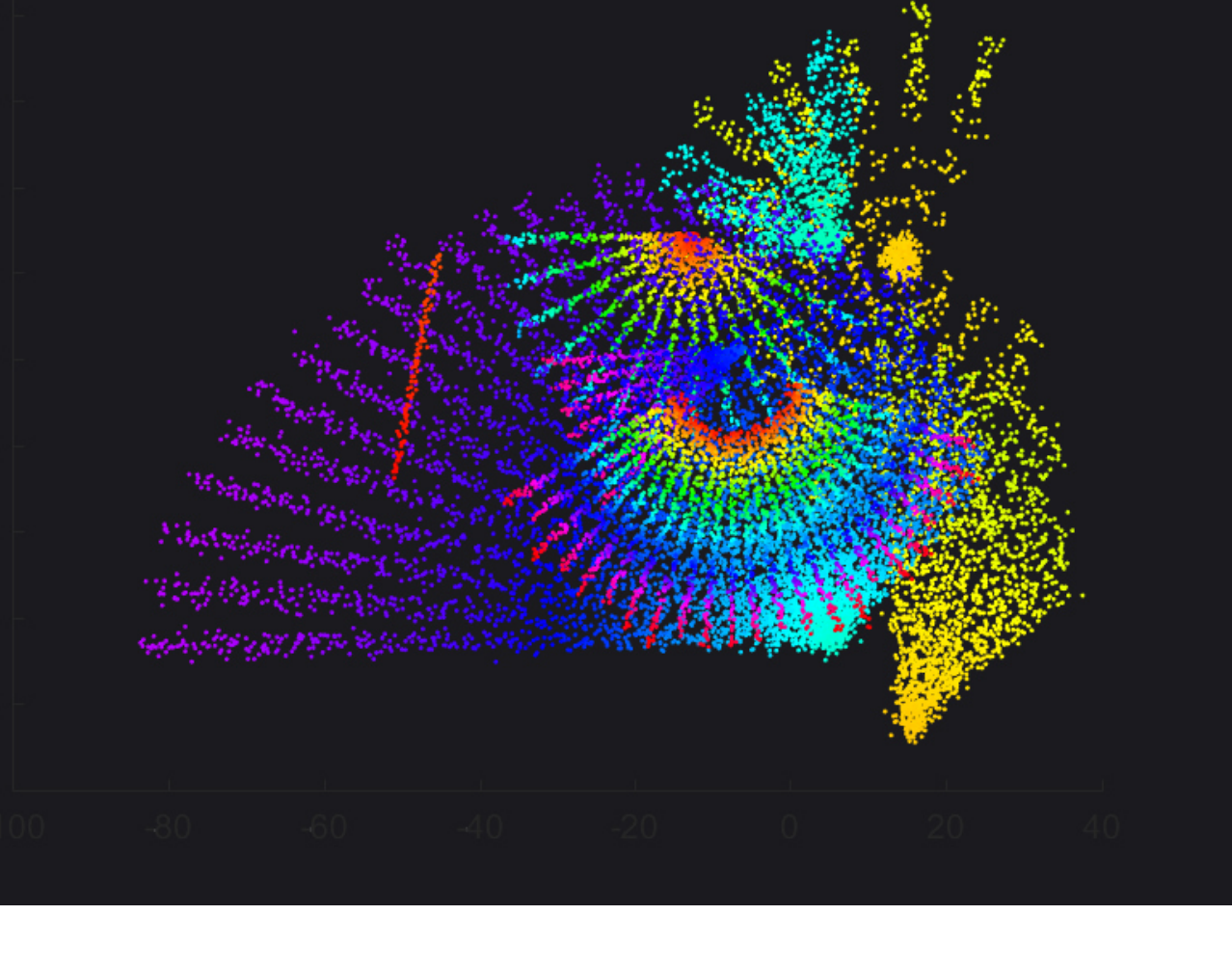}

\includegraphics[trim = 40mm 95mm 40mm 100mm,clip,width=5.8cm, height=4.8cm]{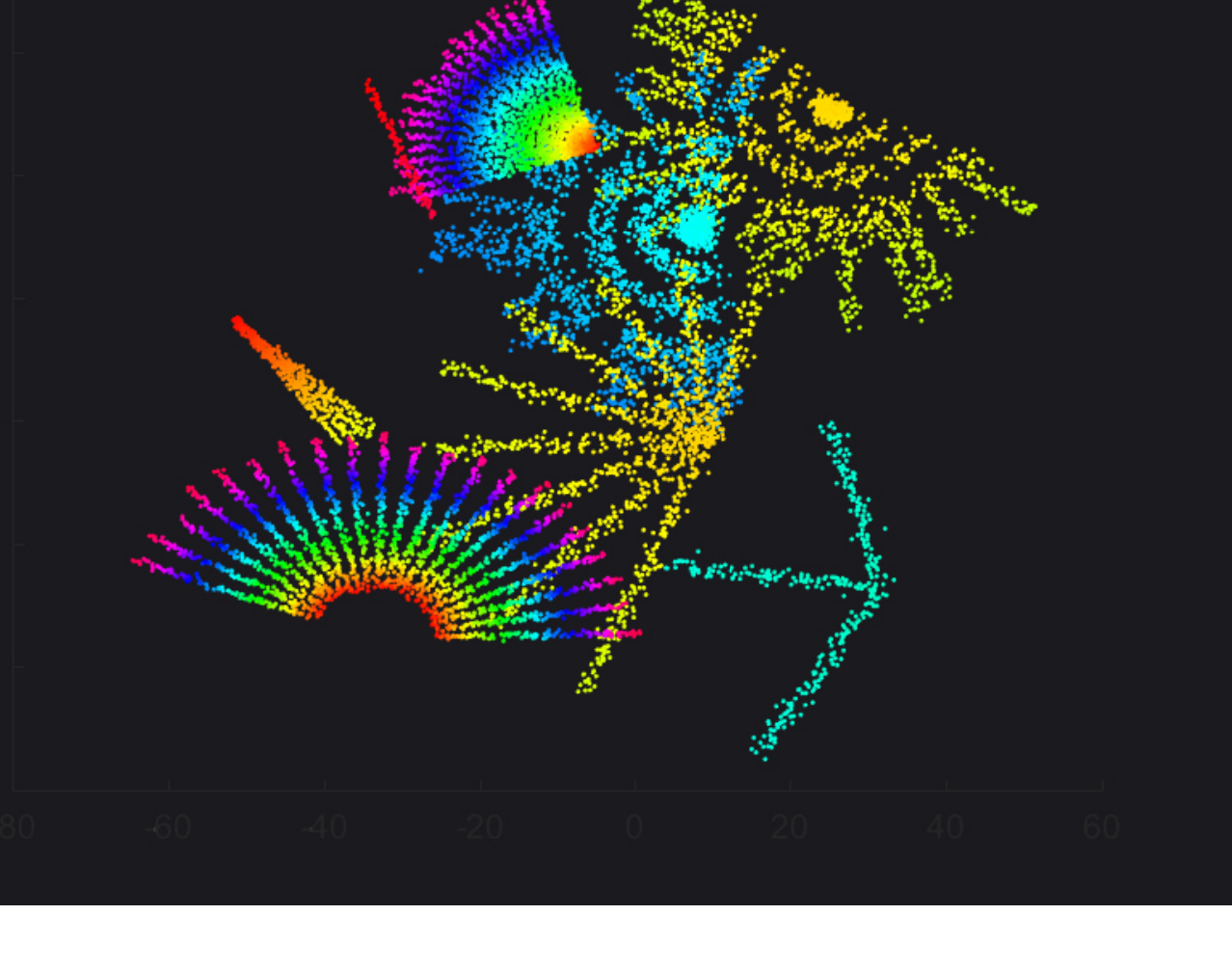}
\includegraphics[trim = 40mm 95mm 40mm 100mm,clip,width=5.8cm, height=4.8cm]{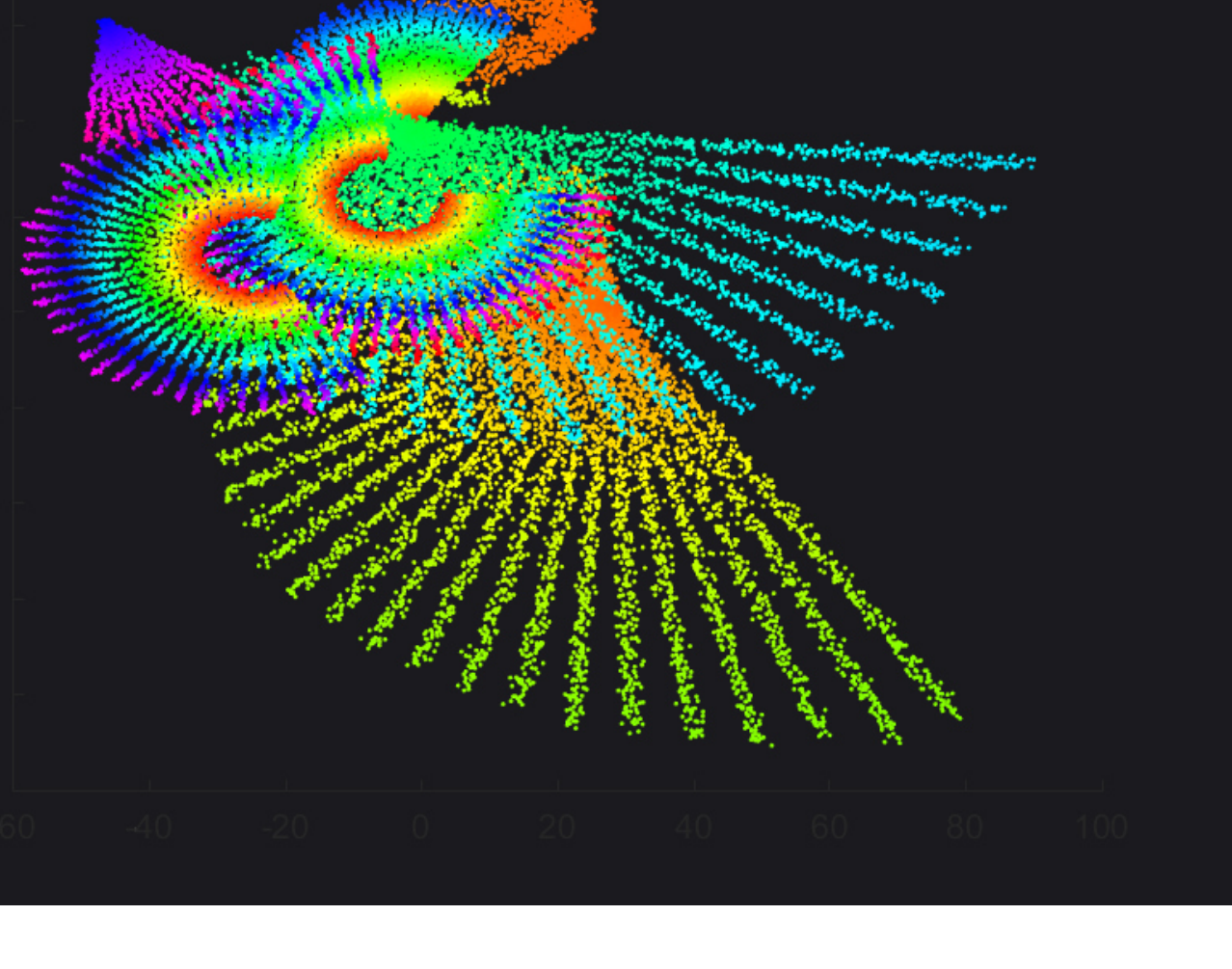}
\includegraphics[trim = 40mm 95mm 40mm 100mm,clip,width=5.8cm, height=4.8cm]{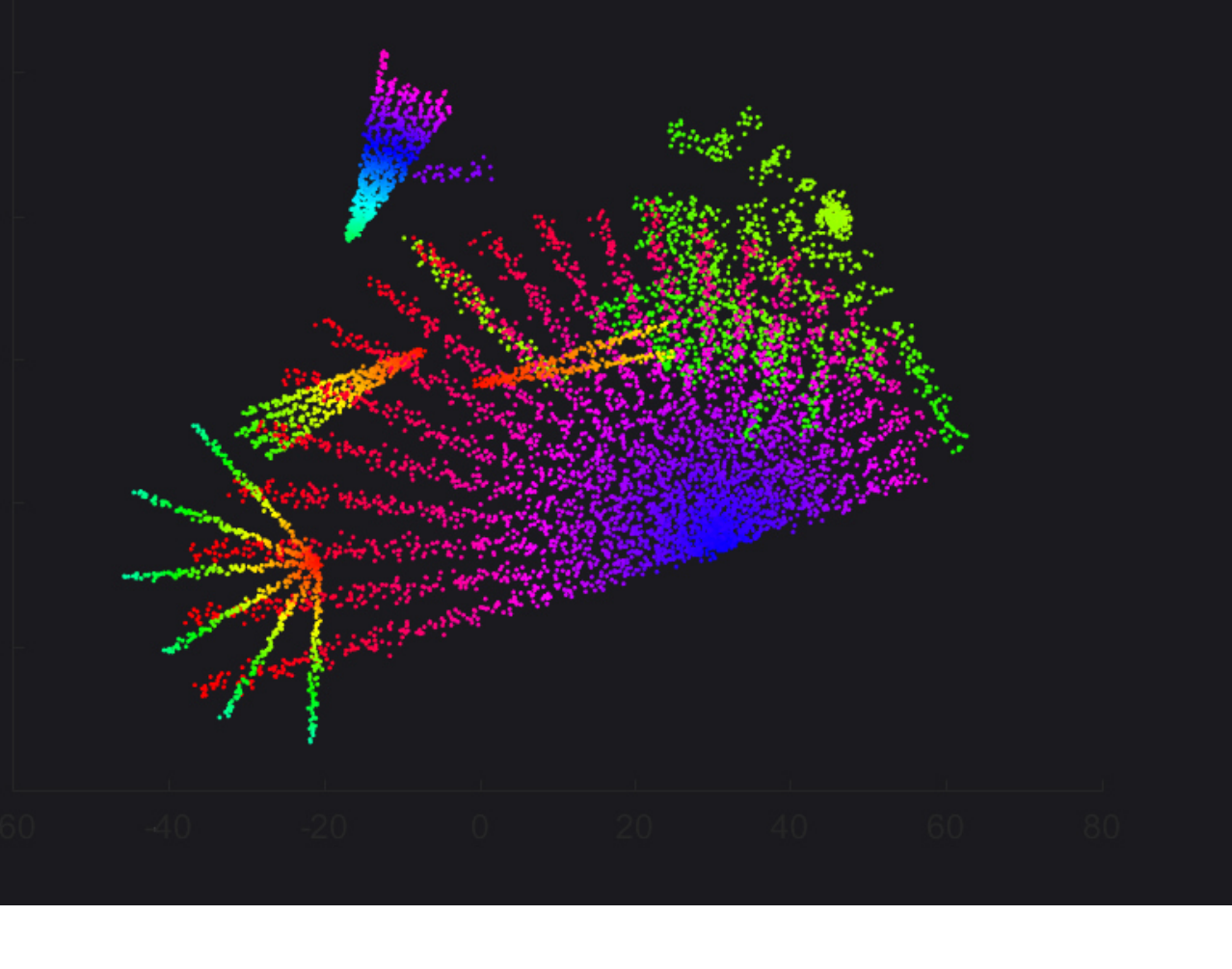}

\caption{Gallery with examples of sand--bubbler inspired art works generated by Algorithm 1.}
\label{fig:examples1}
\end{figure}

Fig. \ref{fig:examples1} shows a gallery of nine art works generated by Algorithm 1 with the parameter values given in Tab.  \ref{tab:design}.  Looking at these images, we can see that they all  display structures that have similarity to natural sand--bubbler patterns. However, colorization of the actual choices of the random--dependent design parameters also offers diversity. 
\begin{figure}[tb]
\includegraphics[trim = 40mm 95mm 40mm 100mm,clip,width=5.8cm, height=4.8cm]{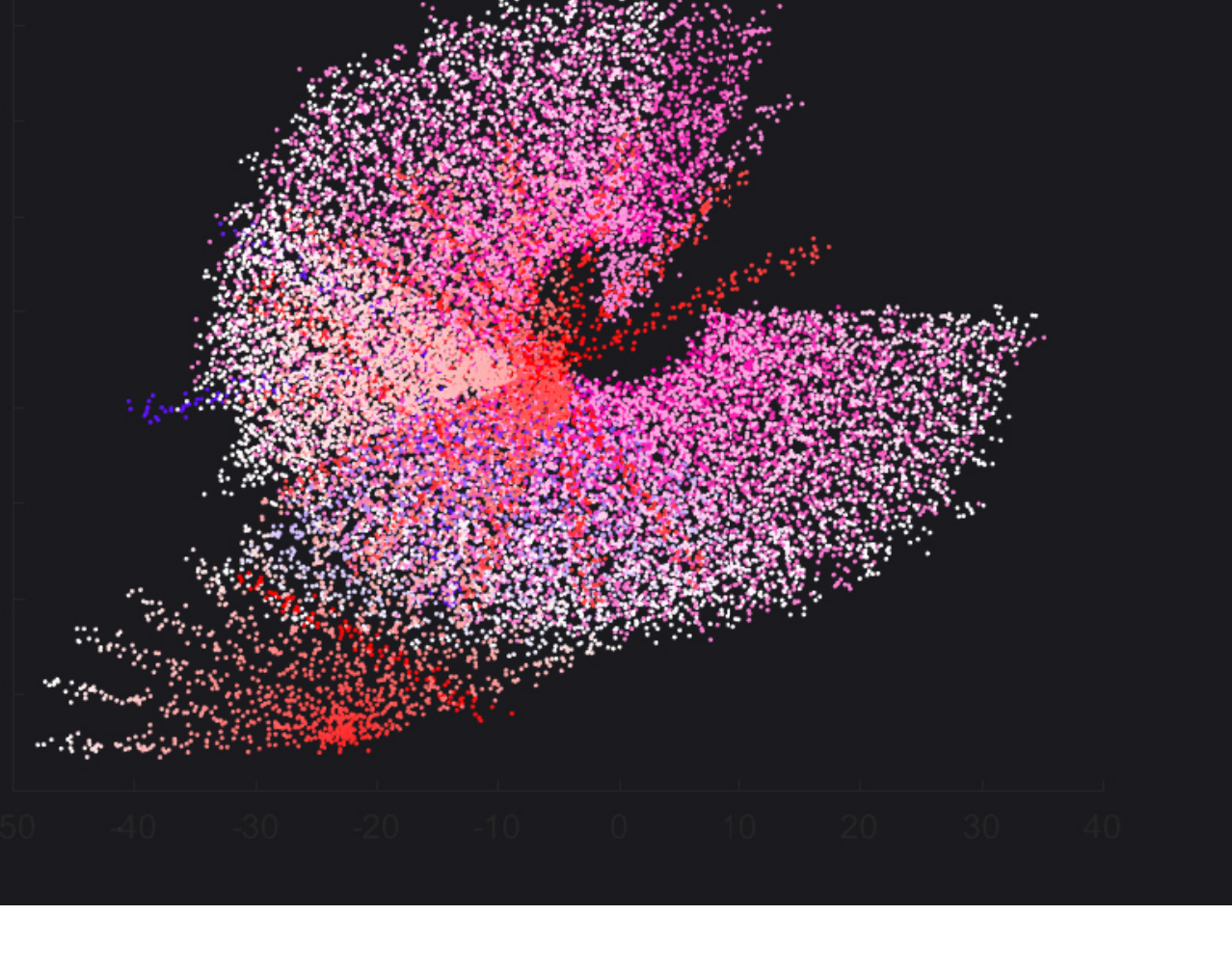}
\includegraphics[trim = 40mm 95mm 40mm 100mm,clip,width=5.8cm, height=4.8cm]{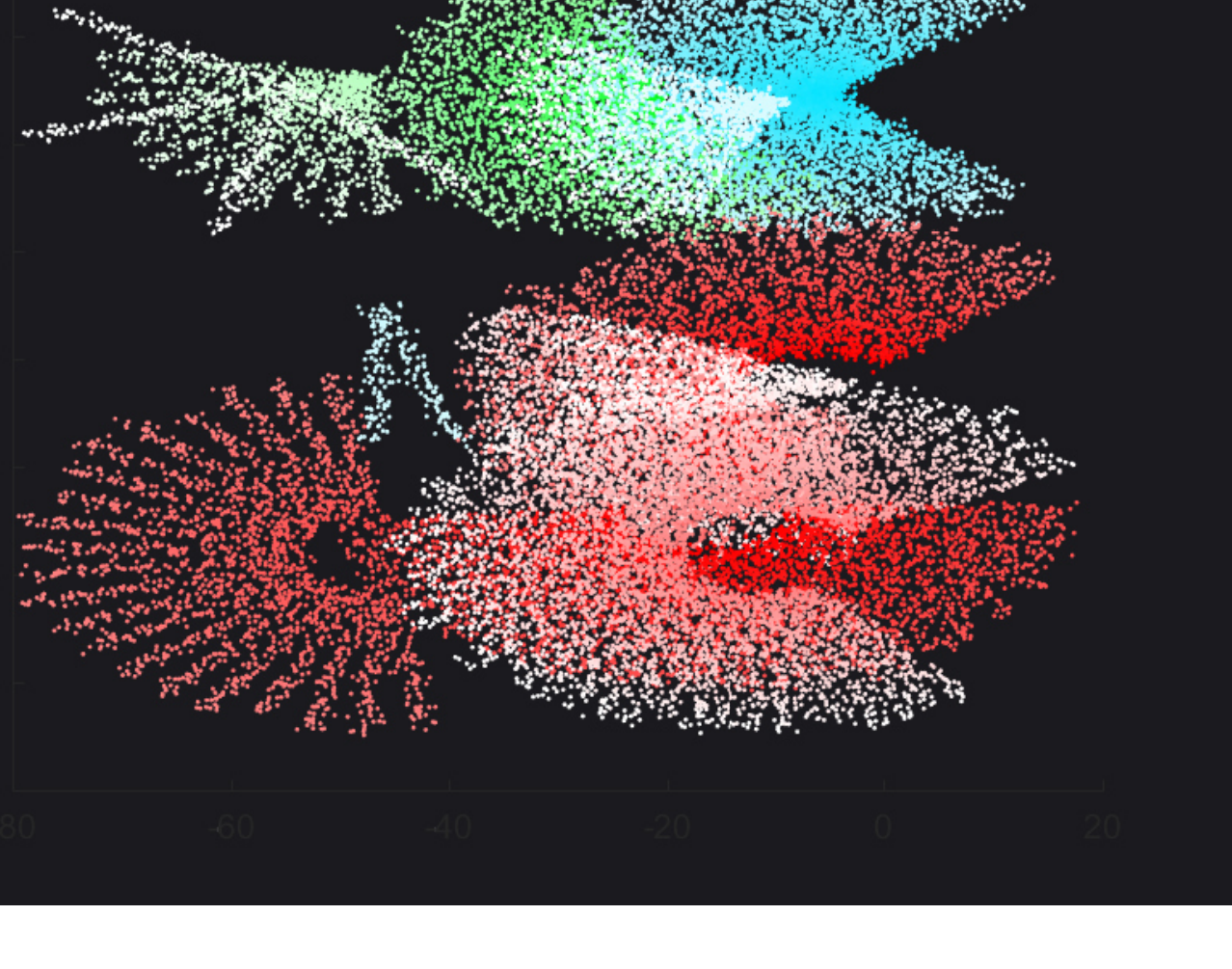}
\includegraphics[trim = 40mm 95mm 40mm 100mm,clip,width=5.8cm, height=4.8cm]{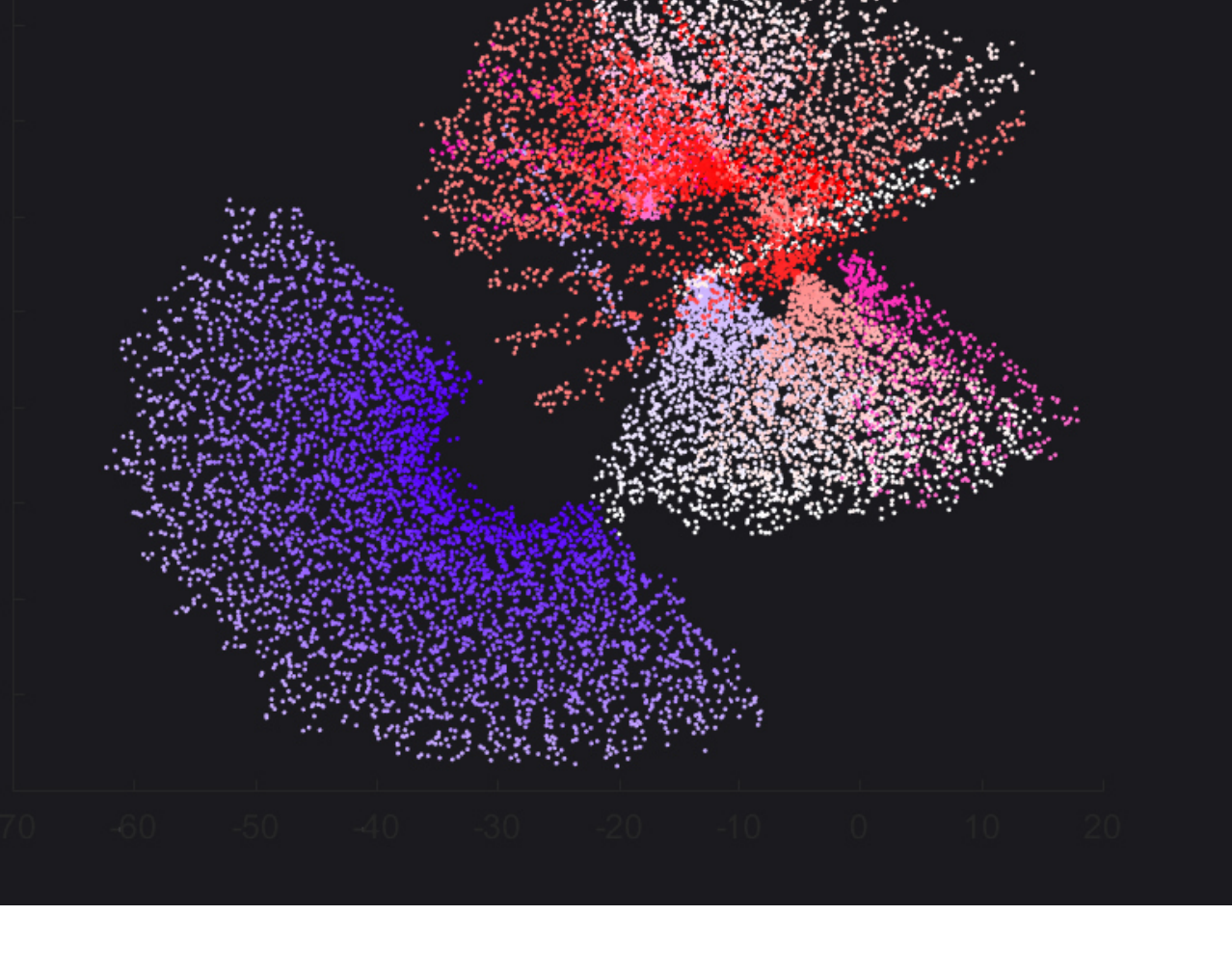}

\includegraphics[trim = 40mm 95mm 40mm 100mm,clip,width=5.8cm, height=4.8cm]{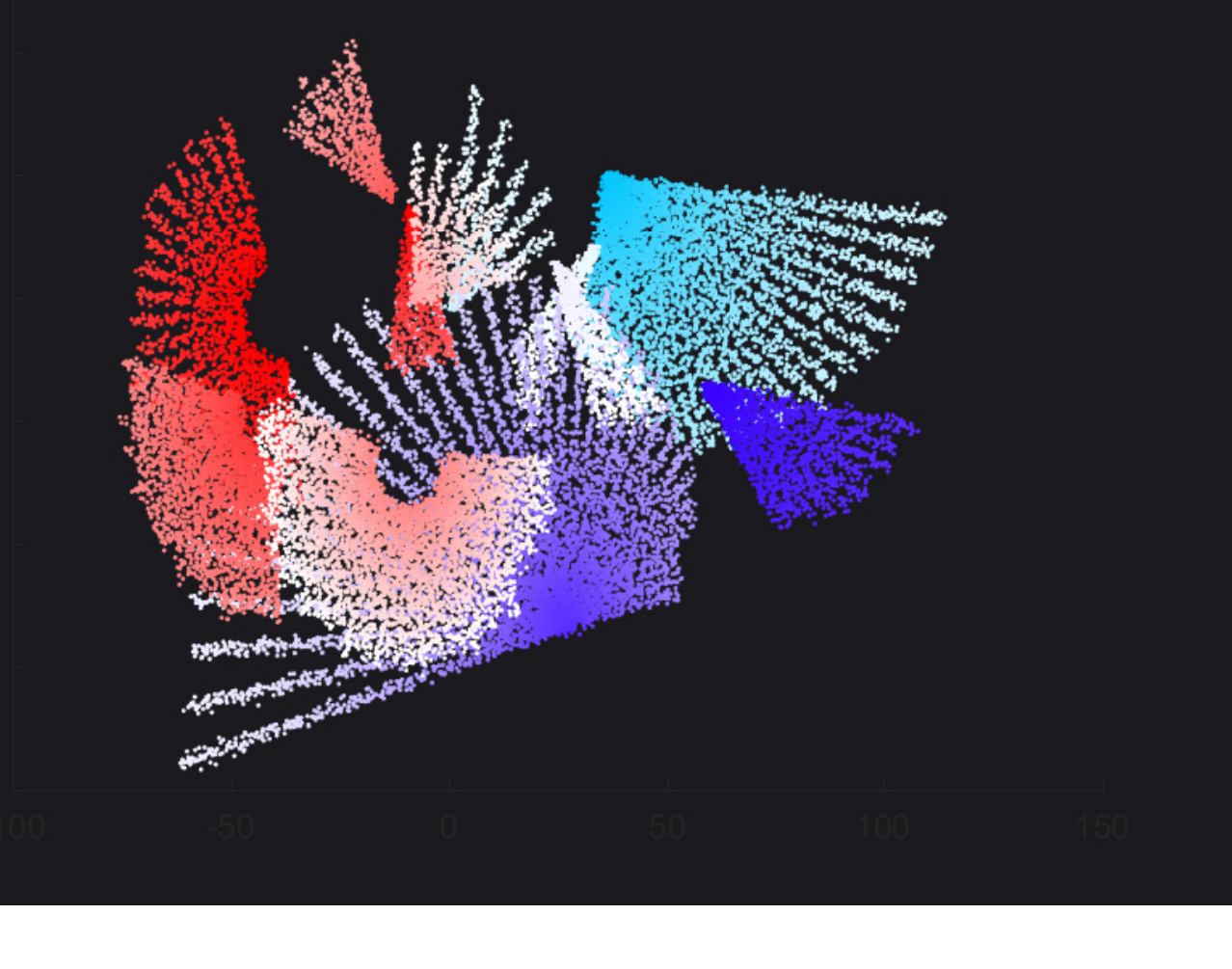}
\includegraphics[trim = 40mm 95mm 40mm 100mm,clip,width=5.8cm, height=4.8cm]{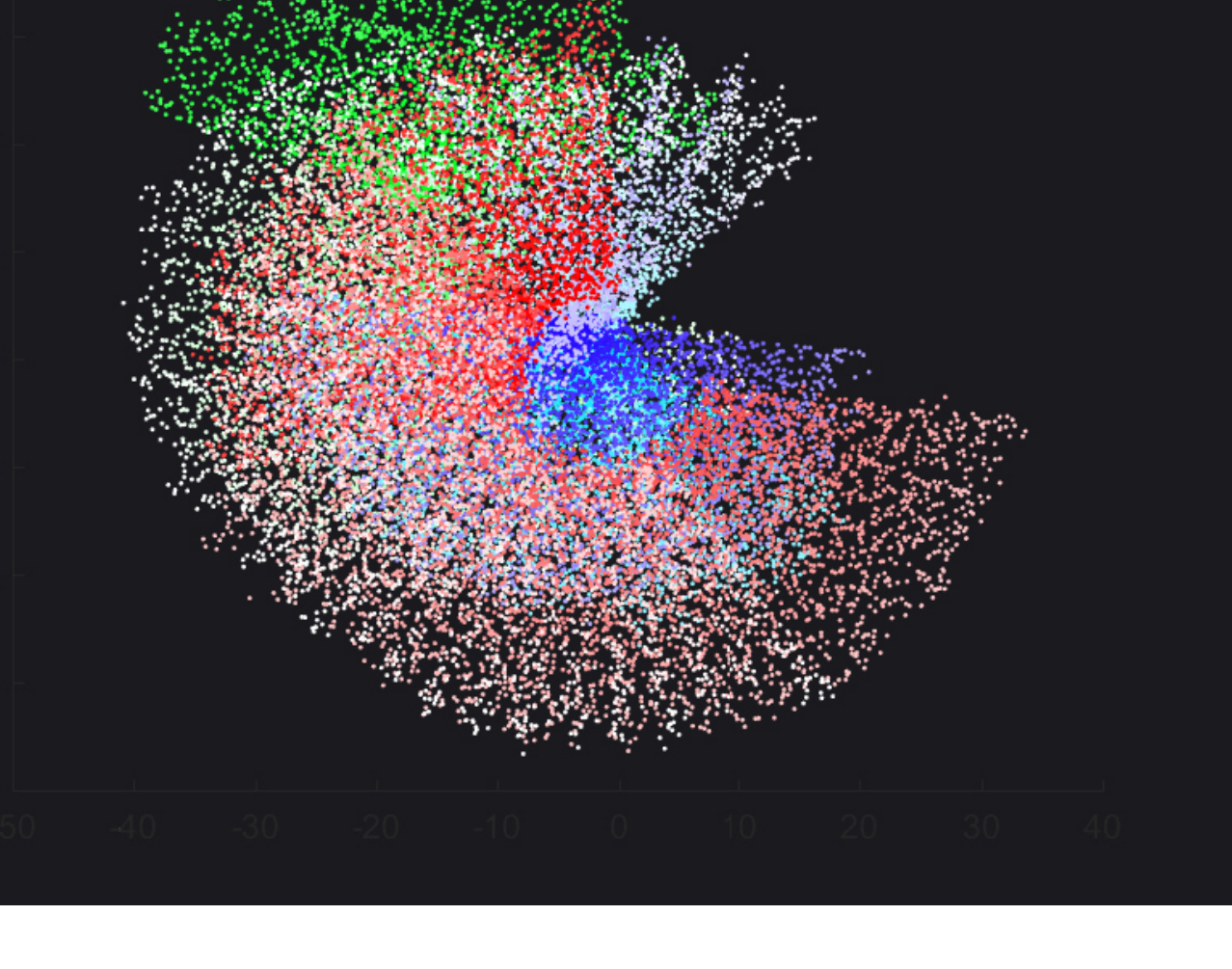}
\includegraphics[trim = 40mm 95mm 40mm 100mm,clip,width=5.8cm, height=4.8cm]{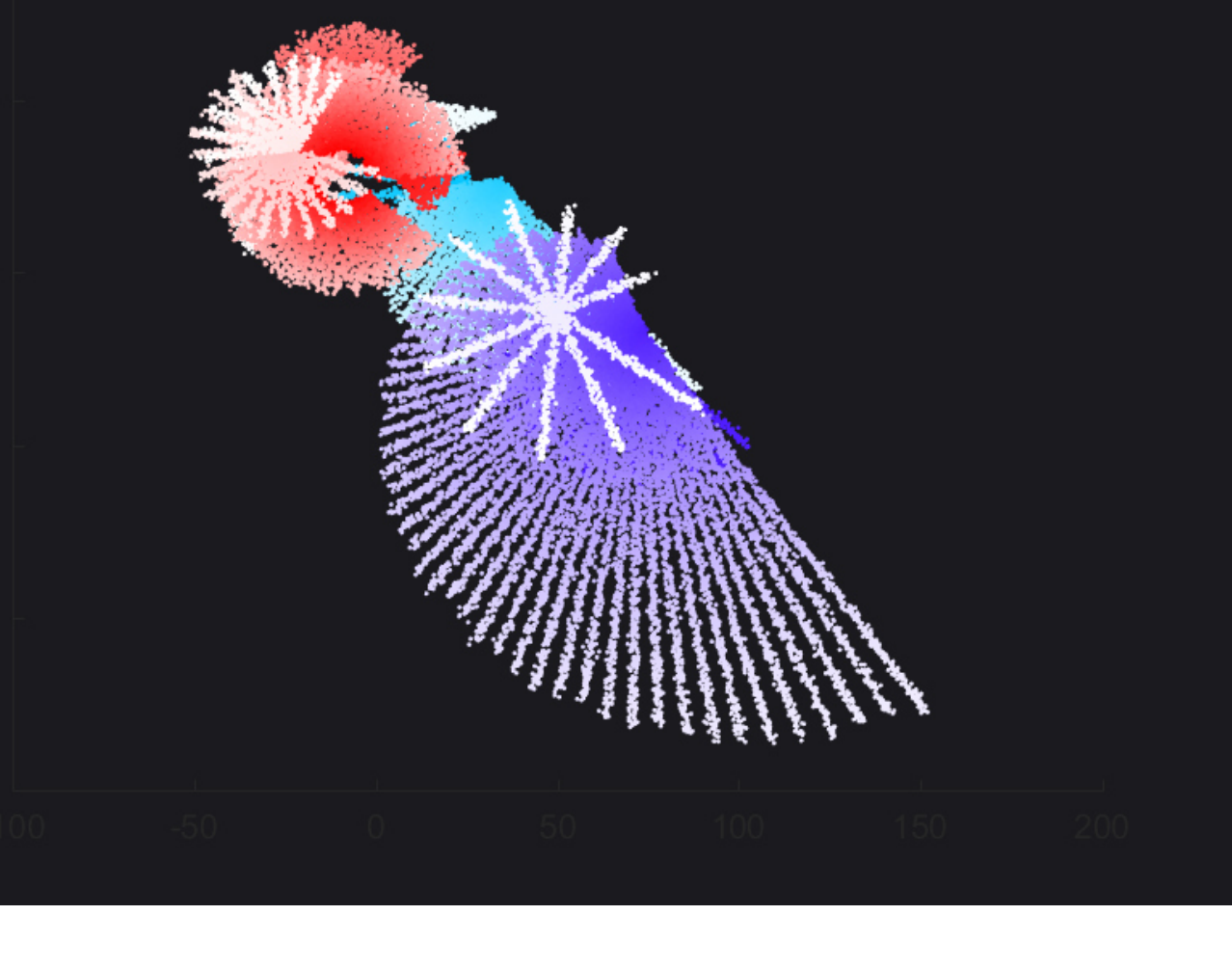}

\caption{Gallery with examples of sand--bubbler inspired art works generated by Algorithm 2.}
\label{fig:examples2}
\end{figure}
In addition, another  six images generated by Algorithm 2 are shown in Fig. \ref{fig:examples2}. Although there is a substantial degree of similarity between these images and the images generated by Algorithm 1, it is worth observing some subtle differences. For instance, it is noticeable that typically the images produced by Algorithm 2 are more widespread and grainier, see for instance Fig. \ref{fig:examples2} upper left and lower middle, which may be attributed to the effect 
of promoting the visual effects associated with the measure BFL. However, at least in the opinion of the author, there was no real success in provoking a distinct color progression or a particularly low colorfulness, as supposedly associated with RRZ and FRD. More analytic work is needed to clarify the relationships between aesthetic measures and visual effects for a given image--producing algorithm.

\section{Concluding remarks and outlook}
An algorithm to create visual art has been  presented that draws inspiration from the collective feeding behavior of sand--bubbler crabs. 
Thus, the paper exemplifies another instance of an art--making process based on mathematical models of animal behavior observed in nature. Apart from the algorithmic design and its artistic output,   
we mainly studied the algorithmic process, particularly how computational aesthetic measures scale to the design parameters and how these measures can be used for controlling and guiding generative evolution.
The interest in analyzing the algorithmic process of digital art--making is from both a computational {\it and} an artistic point of view.  The rationale for the latter is that 
art--theoretically and following an argument of Philip Galanter~\cite{gal10,gal16}, in generative art the artistic work in itself is not seen as important as the artistic process. In a reminiscence to the art movement of ``truth to material'', generative art may focus on ``truth to process''. A possible interpretation states that it is far less interesting to just ask if a human observer likes a particular piece of artificial art. What is much more interesting is to analyze the creation procedure of generative art and study the interplay between algorithmic design (and design parameters) and the works thus created. Not only is the question of whether or not an artificially created image is more or less aesthetic from a human point of view merely opening up  an inexhaustible topic of debate and  might be not answerable, this question may actually not be very useful for advancing our understanding of what  constitutes artistic beauty and how it can be algorithmically (re--)created.   
 In this spirit the galleries of art works shown in Figs. \ref{fig:examples1} and \ref{fig:examples2} could be rather seen as  accompanying the analytic results, and not the other way around.

The analytic and visual results given here only use a small subset of the design space opened up by Algorithm 1 and 2. Future work could further explore this design space. For instance, the images only contain pellets of a given size and color.  It would be interesting to study the visual effects of  
pellets with different sizes  or  pellets decorated in a polychromatic way and/or with their own visual structure. 
Furthermore, the patterns presented are bound by the restriction that they resemble (at least roughly) sand--bubbler patterns as can be observed in
 nature. A possible extension is to soften this restriction. For instance, in nature the scale of patterns may vary, but not dramatically. In the art works, there could be  (next to patterns of the size as seen in Figs. \ref{fig:examples1} and \ref{fig:examples2}) very large structures even exceeding  the edges of the images.  A possible effect would be to  have a background of pellets across the entire image, which may produce different levels of  
pointillism, as similarly shown by Urbano's sand painting artists~\cite{urb11}.

\end{document}